\documentclass[runningheads]{llncs}

\usepackage{eccv}

\usepackage{eccvabbrv}

\usepackage{graphicx}
\usepackage{booktabs}
\usepackage{multirow}

\usepackage[accsupp]{axessibility}  

\usepackage[graphicx]{realboxes}
\usepackage{rotating,booktabs}
\usepackage{hyperref}

\usepackage{orcidlink}

\newcommand*{\Comb}[2]{{}^{#1}C_{#2}}%

\begin{document}

\title{Navigating Text-to-Image Generative Bias across Indic Languages} 


\author{Surbhi Mittal\inst{1}\orcidlink{0000-0001-8910-4161} \and
Arnav Sudan\inst{1} \and
Mayank Vatsa\inst{1}\orcidlink{0000-0001-5952-2274} \and
Richa Singh\inst{1}\orcidlink{0000-0003-4060-4573} \and \\
Tamar Glaser\inst{2}\orcidlink{0009-0007-9322-2444} \and
Tal Hassner\inst{3}\orcidlink{0000-0003-2275-1406}}

\authorrunning{S.~Mittal et al.}

\institute{Department of CSE, IIT Jodhpur, Rajasthan, India \and
Meta, Menlo Park, California, USA\\ \and
Weir P.B.C., Alameda, California, USA}

\maketitle

\begin{abstract}
This research investigates biases in text-to-image (TTI) models for the Indic languages widely spoken across India. It evaluates and compares the generative performance and cultural relevance of leading TTI models in these languages against their performance in English. Using the proposed IndicTTI benchmark, we comprehensively assess the performance of 30 Indic languages with two open-source diffusion models and two commercial generation APIs. The primary objective of this benchmark is to evaluate the support for Indic languages in these models and identify areas needing improvement. Given the linguistic diversity of 30 languages spoken by over 1.4 billion people, this benchmark aims to provide a detailed and insightful analysis of TTI models' effectiveness within the Indic linguistic landscape. The data and code for the IndicTTI benchmark can be accessed at \url{https://iab-rubric.org/resources/other-databases/indictti}.

\keywords{text-to-image generation \and multilingual \and bias}

\end{abstract}

\begin{figure*}[t]
 \centering
   \includegraphics[width=0.85\linewidth]{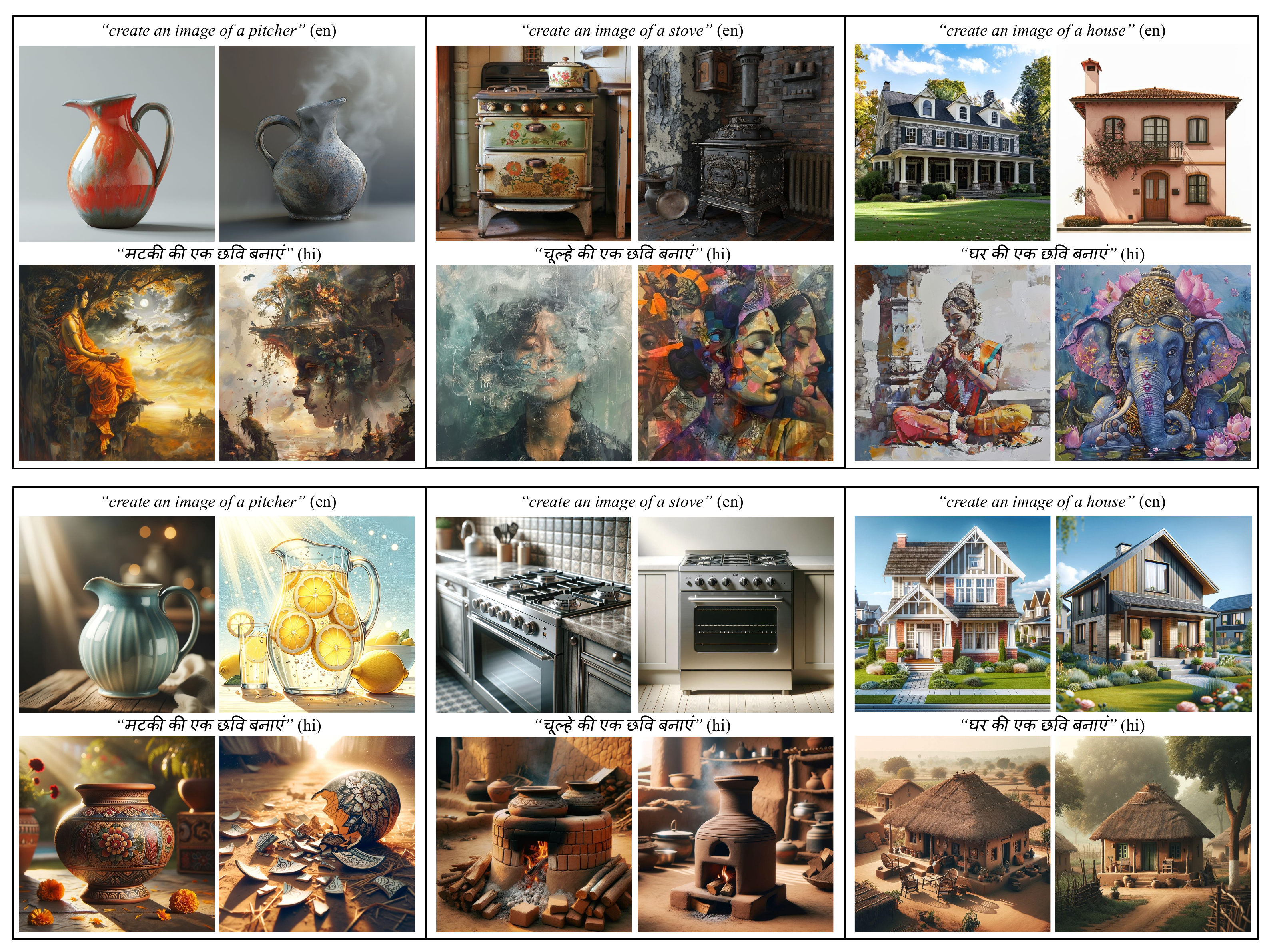}
   \caption{(Top) Images generated by Midjourney when given equivalent prompts in the English and Hindi languages highlighting the tendency of the model to generate incorrectly. (Bottom) Images generated by DallE-3, when given equivalent prompts in the English and Hindi languages, highlight astonishingly different cultural representations.}
   \label{fig:vizabs}
\end{figure*}

\section{Introduction}
\label{sec:intro}

Text-to-image (TTI) generative technologies have transformed the landscape of digital media, distinguished by their extraordinary ability to produce detailed and visually engaging images from written text \cite{rombach2021highresolution, dalle3}. These models have found widespread application in a variety of fields, including gaming, animation, architecture, and fashion, offering significant benefits in terms of enhancing creativity and streamlining production timelines. However, despite the expansion of access through open-source models, challenges persist in achieving linguistic inclusivity \cite{saxon2023multilingual}. The datasets these models are trained on are predominantly composed of English-language content, derived from both proprietary sources\cite{dalle3} and internet scraping \cite{rombach2021highresolution, laion}. This reliance on English-centric data adversely affects the quality of images generated from text in languages other than English.

As the integration of TTI models into various industries deepens, it becomes imperative to conduct a thorough evaluation of their performance and the diversity of the outputs they produce. While there has been research to study biases in TTI models\cite{chen2024would, friedrich2024multilingual}, the current evaluation landscape is notably lacking in benchmarks capable of effectively measuring these models' performance across a diverse linguistic spectrum, often overlooking the rich variety of global languages  \cite{saxon2023multilingual}. To address this gap, we introduce the \textit{IndicTTI benchmark}, an evaluation benchmark designed to explore and measure the biases present in text-to-image generation technologies, with an emphasis on a wider range of languages beyond those typically studied. This benchmark seeks to explore the linguistic diversity of a global population exceeding 1.4 billion, aiming not just to spotlight the shortcomings of existing TTI models but also to foster a broader application of these technologies that is truly inclusive and reflective of the myriad languages and cultures worldwide. Through this in-depth analysis, our goal is to spur improvements in TTI technologies, ensuring they cater to a wider audience and promote a more inclusive environment within digital media creation. 

In this research, we examine the performance of existing models in 30 languages other than English, with a specific focus on Indic languages, which are spoken by over a billion people worldwide. We believe that this benchmark is a step toward analyzing TTI models for a rich and diverse culture with a large population \cite{indianlangs}. The 30 languages are written in 10 different scripts and belong to multiple language families. Among these, we have included the Scheduled Languages, which are considered the major literary languages of India and have a significant volume of literature \cite{indianlangs}. 
We study two aspects of TTI models for assessing bias towards Indic languages, (i) model performance and (ii) model representation. We identify issues of correctness in generation, the presence of cultural elements, and the differences between open-source and API-based models. We utilize Stable Diffusion \cite{rombach2021highresolution} and Alt Diffusion \cite{ye2023altdiffusion} models as part of our open-source analysis, and Midjourney \cite{midjourney} and Dalle3 \cite{ramesh2021zero, dalle3} as part of the API-based generative models. We propose the Cyclic Language-Grounded Correctness (CLGC) metric to evaluate model performance and the Self-Consistency Across Languages (SCAL) metric to evaluate representation bias in the model generations. We further evaluate the model performance through Language-Grounded Correctness (LGC) and Image-Grounded Correctness (IGC) metrics.  Through our analysis, we observe how certain language scripts evoke imagery of Indian Gods in Midjourney, the overwhelmingly high generation of Asian women and couples in open-source models, and the high accuracy of generations in Dalle3 (Refer Fig. \ref{fig:vizabs}). Finally, we discuss the correlation between language script and cultural influences observed in the models.

\begin{table}[t]
\centering
\tiny
\caption{\label{tab:languages} The list of Indic languages included in the IndicTTI benchmark. \textsuperscript{\textdagger}These languages are present in more than one script in the benchmark.
*Languages not listed as part of Indic languages as per the Indian Constitution but spoken by millions.}
\begin{tabular}{|llllll|}
\hline
Assamese   & Malayalam  & Bodo       & Marathi                                                         & Bengali    & Manipuri\textsuperscript\textdagger \\
asm (Beng) & mal (Mlym) & brx (Deva) & mar (Deva)                                                      & ben (Beng) & mni (Beng, Mtei)                                                  \\ \hline
Odia       & Konkani    & Punjabi    & Tamil                                                           & Sanskrit   & Hindi                                                             \\
ory (Orya) & gom (Deva) & pan (Guru) & tam (Taml)                                                      & san (Deva) & hin (Deva)                                                        \\ \hline
Telugu     & Maithili   & Urdu       & Kannada                                                         & Gujarati   & Kashmiri\textsuperscript\textdagger \\
tel (Telu) & mai (Deva) & urd (Arab) & kan (Knda)                                                      & guj (Gujr) & kas (Arab, Deva)                                                  \\ \hline
Awadhi*    & Bhojpuri*  & Magahi*    & Sinhala*                                                        & \multicolumn{2}{l|}{Chhattisgarhi*}                                            \\
awa (Deva) & bho (Deva) & mag (Deva) & sin (Sinh)                                                      & \multicolumn{2}{l|}{hne (Deva)}                                                \\ \hline
Dogri      & Nepali     & Santali    & Sindhi\textsuperscript\textdagger &            &                                                                   \\
doi (Deva) & npi (Deva) & sat (Olck) & snd (Deva, Arab)                                                &            &                                                                   \\ \hline
\end{tabular}
\end{table}

\section{Related Work}

Limited studies have been conducted to benchmark the performance of TTI models across different languages\cite{saxon2023multilingual, liu2023cultural}. In NLP-based tasks, MEGA \cite{ahuja2023mega} benchmarks generative Large Language Models (LLMs) across 16 NLP datasets covering 70 diverse languages. This evaluation compares the performance of generative LLMs like Chat-GPT and GPT-4. Similarly, with a specific focus on the Indic languages, in IndicTrans2\cite{gala2023indictrans}, the authors provide a transformer-based multilingual NMT model that supports high-quality translations across all the 22 scheduled Indic languages. 

In the domain of text-to-image generative models, the recent Coco-crola benchmark \cite{saxon2023multilingual} focuses on the evaluation of DallE and Stable Diffusion models for seven languages, namely English, Spanish, German, Chinese, Japanese, Hebrew, and Indonesian. Among these languages, English, German, and Indonesian share the Latin script, whereas Japanese and Chinese use the same script. Further, the authors focus on the generation of simple concepts through translation; however, in everyday applications, users are more likely to generate images through complex prompts. With this in mind, we focus on more complex prompts describing multiple elements in the generated image. The authors also propose the distinctiveness, coverage, and self-consistency measures.  Very limited research has also been conducted on the design of multilingual text-to-image generative models \cite{li2023translation, ye2023altdiffusion}. One such model is the  AltDiffusion model, which is a multilingual TTI diffusion model designed to process multiple languages \cite{ye2023altdiffusion}. Recently, Struppek et al. \cite{struppek2023exploiting} have shown the influence of script-specific characters on the cultural aspects of the generated images.

In this work, we study the correctness and cultural representation bias of the TTI models, with a focus on Indic languages. However, the metrics and analysis in this study can be organically extended for the analysis of more languages.

\begin{figure*}[t]
 \centering
   \includegraphics[width=0.85\linewidth]{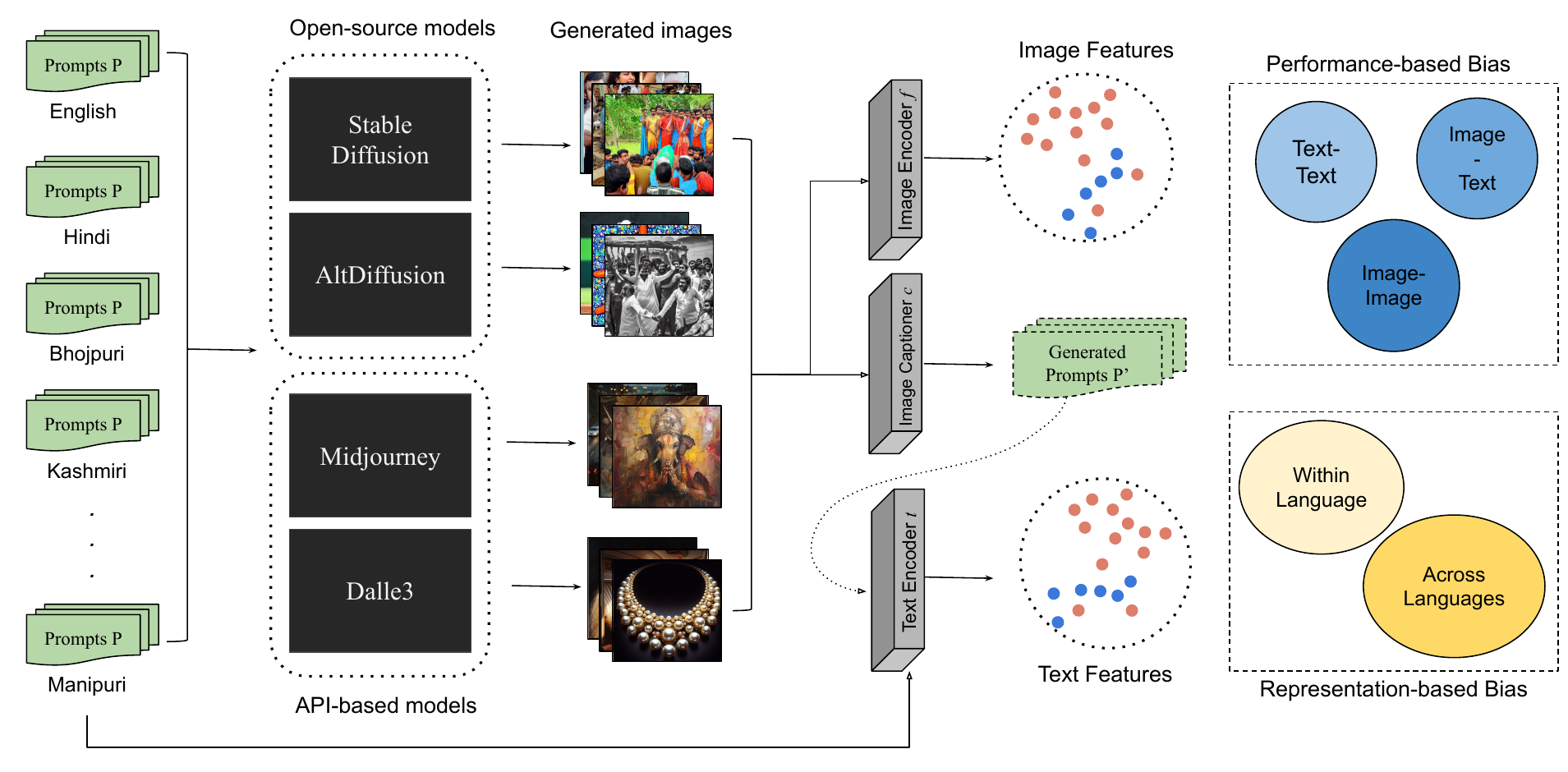}
   \caption{Pipeline of the generation and evaluation of the IndicTTI benchmark. After generating images from four TTI models, their bias is measured across the parameters of their correctness performance as well as representational diversity. The metrics are computed through the use of high-level semantic features extracted from the generated images and their prompts.}
   \label{fig:pipeline}
\end{figure*}

\section{Benchmark Design}
In this work, we propose the IndicTTI benchmark, which comprises of 31 languages, evaluated with four text-to-image generative models with six evaluation metrics covering aspects of correctness and representation. We describe the various design components of the benchmark below.

\subsection{Indic Languages and Prompts}
For this benchmark, we utilize 30 Indic languages in addition to the English language. The languages are listed in Table \ref{tab:languages}. For prompts, we utilize the COCO-NLLB dataset \cite{coconllb, visheratin2023nllb}, which contains image-caption pairs for over 500K images, along with captions translated into 200 languages. In previous work\cite{saxon2023multilingual}, the authors use simplistic concepts for prompts such as ``a photo of a plane." However, the expectation from these models is to yield accurate and unbiased results in day-to-day usage. We believe that this expectation is better met with more complex prompts such as those present in the COCO-NLLB dataset. Therefore, we sample 1000 diverse image-caption pairs from the dataset. The dataset contains 24 Indic languages for which the captions are directly taken. These 24 languages do not include 4 languages that form a part of the 22 scheduled languages of India as per the Indian Constitution, namely Bodo, Dogri,  Konkani, and Santali. Using the IndicTrans2 translator \cite{gala2023indictrans}, we translate English captions to these languages along with two other languages, Manipuri (in Meitei script) and Sindhi (in Devanagari script). This leads to a total of 1000 prompts in 31 languages, including English, along with an associated image. 

\subsection{TTI Models and Generated Images}
We utilize four different text-to-image models for the benchmark. For \textbf{open-source models}, we use the Stable Diffusion \cite{rombach2021highresolution} and AltDiffusion Models\cite{ye2023altdiffusion}. For \textbf{API-based models}, we use Dalle3\cite{dalle3} and Midjourney\cite{midjourney}.  

Using the open-source models, we generate 4 images per prompt for the selected 1000 prompts across 31 languages, leading to 124K images corresponding to Stable Diffusion and AltDiffusion each. For API-based models, we generate over a subset of 200 prompts for Midjourney and over 50 prompts for Dalle3. This leads to a total of 24.8K images for Midjourney and 6.2K images for Dalle3. The evaluation is done over these sets and additionally over a subset of 40 prompts common across all models. The Dalle3 and Midjourney APIs are paid, and further details about the generation process are provided in the supplementary.

\section{Evaluation Methods}
In the IndicTTI benchmark, we focus on two aspects of TTI model evaluation- correctness and representation. \textit{Correctness} refers to the ability to measure the semantic faithfulness with which the model generates the images for the corresponding prompt. On the other hand, \textit{representation} refers to the measurement of the diversity of generated images within and across the different languages for the given prompts. The complete pipeline of the study is depicted in Fig. \ref{fig:pipeline}.

We adopt the following notation to describe the metrics and the evaluation pipeline. Let the set of languages be $L$. The set of prompts is denoted by $P = \{p_1, p_2,.. p_k\}$, and the corresponding representative images are $Q = \{q_1, q_2,.. q_k\}$. For each prompt in $P$, we generate $n$ images, resulting in a total of $n|P||L|$ images for a given generative model $g$. The set of images generated by $g$ for a prompt $p_k$ in language $l \in L$ is denoted as,
\begin{equation}
    I_{p_k, l} = \{{i_{p_k, l, r}}\}_{r=0}^n 
\end{equation}
where, $i_{p_k, l, r}$ denotes the image generated using $g$ as $g(p_k, l)$. Utilizing this notation, we define the metrics in this section. Across all the metrics, we compute high-level semantic features of images (using feature extractor $f$) and text (using feature extractor $t$) for effective comparison using the similarity function $\phi$. 

\subsection{Correctness-based Metrics}
We employ three correctness-based metrics in our study, including the proposed Cyclic Language-Grounded Correctness metric. The image-grounded and language-grounded correctness metrics are inspired by the \textit{correctness} metric in CocoCrola\cite{saxon2023multilingual}. These three metrics comprehensively evaluate the generated images across text-text, image-image, and image-text spaces. A higher correctness value indicates greater accuracy of the generative model in producing images that are faithful to the text.

\noindent \textbf{Cyclic Language-Grounded Correctness (CLGC):} This metric evaluates the generated images $I$ in the text-text space. This is achieved in a cyclic manner through generating captions over $I$ using a caption generator $c$, as,
\begin{equation}
    p'_k = c(i_{p_k, l, r})
\end{equation}
where, $p'_k$ denotes the generated caption for $r_{th}$ image in set $I_{p_k,l}$. Then, with $p_k$ and $p'_k$ as prompts in the English language for the language $l$, we obtain,
\begin{equation}
    CLGC(l) = \frac{1}{n|P|} \sum_{p_k}^{P} \sum_{r=0}^{n} \phi [\; t(p_k),\; t(p'_k) \;]
\end{equation}

\noindent \textbf{Image-Grounded Correctness (IGC):} For this metric, the correctness of the generated images is evaluated in the image-image space, as,
\begin{equation}
    IGC(l) = \frac{1}{n|P|} \sum_{p_k}^{P} \sum_{r=0}^{n} \phi [\; f(q_k),\; f(i_{p_k}) \;]
\end{equation}
where, the ground-truth image $q_k$ is compared to each of the $n$ generated images for the prompt $p_k$.

\noindent \textbf{Language-Grounded Correctness (LGC):} This metric evaluates correctness in the image-text space with $p_k$ denoting the prompt in the English language,
\begin{equation}
    LGC(l) = \frac{1}{n|P|} \sum_{p_k}^{P} \sum_{r=0}^{n} \phi [\; t(p_k),\; f(i_{p_k}) \;]
\end{equation}

\subsection{Representation-based Metrics}
In this section, we discuss the three metrics utilized for measuring the representativeness of concepts across the languages. We propose the Self-Consistency Across Languages metric to capture the \textit{bias} in representation across the different languages. The other two metrics are inspired by the \textit{self-consistency} and \textit{distinctiveness} metrics in previous work\cite{saxon2023multilingual}.

\noindent \textbf{Self-Consistency Across Languages (SCAL):} This metric captures the variation in the generated images for the same prompt across the different languages. So, for a language pair $(l_a, l_b)$,
\begin{align}
    SCAL(l_a, l_b, p_k) &= \frac{1}{\Comb{n}{2}} \sum_{u=0}^{n} \sum_{v=0}^{n} \phi [\; f(i_{p_k, l_a, u}),\; f(i_{p_k, l_b, v}) \;]; u \ne v \\
    SCAL(l_a, l_b) &= \frac{1}{|P|}  \sum_{p_k}^{P} SCAL(l_a, l_b, p_k) \\
\end{align}
Then, the overall SCAL score for the given set of languages becomes,
\begin{equation}
\label{eqn:scal}
    SCAL = \frac{1}{\Comb{|L|}{2}} \sum_{l_a}^{L} \sum_{l_b}^{L} SCAL(l_a, l_b);\; a \ne b
\end{equation}
The lower the value of SCAL, the less consistent the generations are across the different languages. On the other hand, a higher value of SCAL would demonstrate high consistency between concepts across the different languages.

\noindent \textbf{Self-Consistency Within Language (SCWL):} This metric is computed within the images generated for a particular prompt for a given language, as,
\begin{align}
    SCWL(l, p_k) &= \frac{1}{\Comb{n}{2}} \sum_{u=0}^{n} \sum_{v=0}^{n} \phi [\; f(i_{p_k, u}),\; f(i_{p_k, v}) \;]; u \ne v \\
    SCWL(l) &= \frac{1}{|P|}  \sum_{p_k}^{P} SCWL(l, p_k)
\end{align}
A high self-consistency showcases good consistency between the semantic content of the generated images.

\noindent \textbf{Distinctiveness Within Language (DWL):} This metric depicts the diversity across the images generated for the different prompts $p_a$ and $p_b$,
\begin{align}
    SWL(l) &= 1 - \frac{1}{\Comb{|P|}{2}} \sum_{p_a}^{P} \sum_{p_b}^{P} \phi [\; f(i_{p_a}),\; f(i_{p_b}) \;];  a \ne b
\end{align}

where a higher distinctiveness score showcases the capability of the model to generate diverse images with varying prompts.

In addition to quantitative evaluation, we conduct an extensive qualitative evaluation of the generated images with respect to the captions, languages, and scripts. The implementation details are provided in the supplementary.


\begin{sidewaystable}
\tiny
\caption{Cyclic Language Grounded Correctness (CLGC) (\%) across the different Indic languages in the IndicTTI benchmark. Existing models provide high correctness for English languages while providing lower values for Indic languages. } \label{tab:clgc40}
\centering
\begin{tabular}{|l|c|ccccccccccccccc|}
\hline
Model            & en    & \begin{tabular}[c]{@{}c@{}}asm \\ (Beng)\end{tabular} & \begin{tabular}[c]{@{}c@{}}ben\\ (Beng)\end{tabular} & \begin{tabular}[c]{@{}c@{}}guj\\ (Gujr)\end{tabular} & \begin{tabular}[c]{@{}c@{}}hin\\ (Deva)\end{tabular} & \begin{tabular}[c]{@{}c@{}}kan\\ (Knda)\end{tabular} & \begin{tabular}[c]{@{}c@{}}mal\\ (Mlym)\end{tabular} & \begin{tabular}[c]{@{}c@{}}mar\\ (Deva)\end{tabular} & \begin{tabular}[c]{@{}c@{}}mni\\ (Beng)\end{tabular} & \begin{tabular}[c]{@{}c@{}}npi\\ (Deva)\end{tabular} & \begin{tabular}[c]{@{}c@{}}ory\\ (Orya)\end{tabular} & \begin{tabular}[c]{@{}c@{}}pan \\ (Guru)\end{tabular} & \begin{tabular}[c]{@{}c@{}}san\\ (Deva)\end{tabular} & \begin{tabular}[c]{@{}c@{}}snd\\ (Arab)\end{tabular} & \begin{tabular}[c]{@{}c@{}}tam\\ (Taml)\end{tabular} & \begin{tabular}[c]{@{}c@{}}tel\\ (Telu)\end{tabular} \\ \hline
Stable Diffusion & 64.53 & 7.29                                                  & 6.39                                                 & 7.24                                                 & 7.35                                                 & 7.13                                                 & 6.91                                                 & 7.53                                                 & 6.50                                                 & 8.03                                                 & 8.36                                                 & 8.78                                                  & 8.14                                                 & 11.29                                                & 7.30                                                 & 7.65                                                 \\
Alt Diffusion    & 58.16 & 14.65                                                 & 16.70                                                & 15.97                                                & 16.79                                                & 22.98                                                & 16.08                                                & 16.50                                                & 12.17                                                & 17.97                                                & 18.82                                                & 17.64                                                 & 17.99                                                & 14.82                                                & 18.10                                                & 14.50                                                \\
Midjourney       & 64.31 & 12.99                                                 & 11.89                                                & 12.49                                                & 11.31                                                & 12.12                                                & 10.92                                                & 11.99                                                & 9.80                                                 & 11.89                                                & 12.56                                                & 9.32                                                  & 12.19                                                & 13.49                                                & 12.84                                                & 12.16                                                \\
Dalle3           & 65.70 & 53.36                                                 & 60.73                                                & 57.25                                                & 64.16                                                & 60.13                                                & 57.68                                                & 61.84                                                & 35.24                                                & 59.90                                                & 53.96                                                & 62.67                                                 & 51.22                                                & 45.19                                                & 53.98                                                & 55.26                                                \\ \hline
Model            & en    & \begin{tabular}[c]{@{}c@{}}urd\\ (Arab)\end{tabular}  & \begin{tabular}[c]{@{}c@{}}kas\\ (Arab)\end{tabular} & \begin{tabular}[c]{@{}c@{}}kas\\ (Deva)\end{tabular} & \begin{tabular}[c]{@{}c@{}}mai\\ (Deva)\end{tabular} & \begin{tabular}[c]{@{}c@{}}awa\\ (Deva)\end{tabular} & \begin{tabular}[c]{@{}c@{}}bho\\ (Deva)\end{tabular} & \begin{tabular}[c]{@{}c@{}}hne\\ (Deva)\end{tabular} & \begin{tabular}[c]{@{}c@{}}mag\\ (Deva)\end{tabular} & \begin{tabular}[c]{@{}c@{}}sin\\ (Sinh)\end{tabular} & \begin{tabular}[c]{@{}c@{}}brx\\ (Deva)\end{tabular} & \begin{tabular}[c]{@{}c@{}}doi\\ (Deva)\end{tabular}  & \begin{tabular}[c]{@{}c@{}}gom\\ (Deva)\end{tabular} & \begin{tabular}[c]{@{}c@{}}sat\\ (Olck)\end{tabular} & \begin{tabular}[c]{@{}c@{}}snd\\ (Deva)\end{tabular} & \begin{tabular}[c]{@{}c@{}}mni\\ (Mtei)\end{tabular} \\ \hline
Stable Diffusion & 64.53 & 10.08                                                 & 6.29                                                 & 7.52                                                 & 7.13                                                 & 10.11                                                & 7.12                                                 & 8.25                                                 & 7.66                                                 & 6.25                                                 & 6.26                                                 & 7.95                                                  & 7.78                                                 & 10.51                                                & 7.59                                                 & 10.17                                                \\
Alt Diffusion    & 58.16 & 17.68                                                 & 15.96                                                & 15.46                                                & 16.93                                                & 16.32                                                & 15.41                                                & 16.74                                                & 16.80                                                & 13.04                                                & 11.98                                                & 13.50                                                 & 15.82                                                & 14.13                                                & 13.37                                                & 15.25                                                \\
Midjourney       & 64.31 & 15.38                                                 & 12.72                                                & 12.37                                                & 11.92                                                & 13.04                                                & 11.47                                                & 12.06                                                & 12.53                                                & 11.96                                                & 12.34                                                & 11.77                                                 & 12.59                                                & 11.36                                                & 12.35                                                & 9.72                                                 \\
Dalle3           & 65.70 & 64.02                                                 & 47.45                                                & 55.51                                                & 58.40                                                & 57.37                                                & 60.75                                                & 59.18                                                & 60.42                                                & 49.03                                                & 28.87                                                & 59.84                                                 & 50.18                                                & 10.93                                                & 56.78                                                & 11.19                                                \\ \hline
\end{tabular}


\caption{Image-Grounded Correctness (IGC) (\%) across the different Indic languages in the IndicTTI benchmark.} \label{tab:igc40}
\tiny
\centering
\begin{tabular}{|l|c|ccccccccccccccc|}
\hline
Model            & en    & \begin{tabular}[c]{@{}c@{}}asm \\ (Beng)\end{tabular} & \begin{tabular}[c]{@{}c@{}}ben\\ (Beng)\end{tabular} & \begin{tabular}[c]{@{}c@{}}guj\\ (Gujr)\end{tabular} & \begin{tabular}[c]{@{}c@{}}hin\\ (Deva)\end{tabular} & \begin{tabular}[c]{@{}c@{}}kan\\ (Knda)\end{tabular} & \begin{tabular}[c]{@{}c@{}}mal\\ (Mlym)\end{tabular} & \begin{tabular}[c]{@{}c@{}}mar\\ (Deva)\end{tabular} & \begin{tabular}[c]{@{}c@{}}mni\\ (Beng)\end{tabular} & \begin{tabular}[c]{@{}c@{}}npi\\ (Deva)\end{tabular} & \begin{tabular}[c]{@{}c@{}}ory\\ (Orya)\end{tabular} & \begin{tabular}[c]{@{}c@{}}pan \\ (Guru)\end{tabular} & \begin{tabular}[c]{@{}c@{}}san\\ (Deva)\end{tabular} & \begin{tabular}[c]{@{}c@{}}snd\\ (Arab)\end{tabular} & \begin{tabular}[c]{@{}c@{}}tam\\ (Taml)\end{tabular} & \begin{tabular}[c]{@{}c@{}}tel\\ (Telu)\end{tabular} \\ \hline
Stable Diffusion & 51.21 & 24.19                                                 & 23.84                                                & 23.25                                                & 23.53                                                & 23.24                                                & 22.96                                                & 23.53                                                & 23.58                                                & 23.96                                                & 23.76                                                & 22.97                                                 & 23.11                                                & 23.06                                                & 22.68                                                & 22.61                                                \\
Alt Diffusion    & 46.58 & 24.23                                                 & 25.08                                                & 25.18                                                & 25.43                                                & 27.74                                                & 24.27                                                & 25.42                                                & 22.79                                                & 25.98                                                & 26.27                                                & 25.48                                                 & 25.68                                                & 24.68                                                & 25.06                                                & 22.19                                                \\
Midjourney       & 48.94 & 22.69                                                 & 22.66                                                & 23.06                                                & 22.50                                                & 22.81                                                & 22.04                                                & 22.75                                                & 21.97                                                & 22.29                                                & 22.82                                                & 23.05                                                 & 22.81                                                & 23.34                                                & 22.55                                                & 22.09                                                \\
Dalle3           & 48.92 & 40.02                                                 & 44.22                                                & 43.75                                                & 44.86                                                & 42.74                                                & 43.12                                                & 46.10                                                & 30.91                                                & 43.14                                                & 40.00                                                & 45.06                                                 & 37.66                                                & 36.30                                                & 41.25                                                & 41.39                                                \\ \hline
Model            & en    & \begin{tabular}[c]{@{}c@{}}urd\\ (Arab)\end{tabular}  & \begin{tabular}[c]{@{}c@{}}kas\\ (Arab)\end{tabular} & \begin{tabular}[c]{@{}c@{}}kas\\ (Deva)\end{tabular} & \begin{tabular}[c]{@{}c@{}}mai\\ (Deva)\end{tabular} & \begin{tabular}[c]{@{}c@{}}awa\\ (Deva)\end{tabular} & \begin{tabular}[c]{@{}c@{}}bho\\ (Deva)\end{tabular} & \begin{tabular}[c]{@{}c@{}}hne\\ (Deva)\end{tabular} & \begin{tabular}[c]{@{}c@{}}mag\\ (Deva)\end{tabular} & \begin{tabular}[c]{@{}c@{}}sin\\ (Sinh)\end{tabular} & \begin{tabular}[c]{@{}c@{}}brx\\ (Deva)\end{tabular} & \begin{tabular}[c]{@{}c@{}}doi\\ (Deva)\end{tabular}  & \begin{tabular}[c]{@{}c@{}}gom\\ (Deva)\end{tabular} & \begin{tabular}[c]{@{}c@{}}sat\\ (Olck)\end{tabular} & \begin{tabular}[c]{@{}c@{}}snd\\ (Deva)\end{tabular} & \begin{tabular}[c]{@{}c@{}}mni\\ (Mtei)\end{tabular} \\ \hline
Stable Diffusion & 51.21 & 22.92                                                 & 22.58                                                & 23.41                                                & 23.71                                                & 25.16                                                & 22.98                                                & 23.19                                                & 23.41                                                & 24.00                                                & 23.83                                                & 22.90                                                 & 23.88                                                & 24.14                                                & 23.64                                                & 23.98                                                \\
Alt Diffusion    & 46.58 & 25.40                                                 & 24.45                                                & 25.01                                                & 26.39                                                & 25.97                                                & 25.53                                                & 25.59                                                & 26.10                                                & 24.60                                                & 23.82                                                & 24.87                                                 & 24.73                                                & 24.16                                                & 25.16                                                & 24.35                                                \\
Midjourney       & 48.94 & 23.30                                                 & 22.76                                                & 22.63                                                & 22.90                                                & 23.09                                                & 22.65                                                & 22.97                                                & 22.64                                                & 22.16                                                & 22.57                                                & 22.90                                                 & 22.79                                                & 21.25                                                & 22.53                                                & 21.99                                                \\
Dalle3           & 48.92 & 44.10                                                 & 38.81                                                & 44.04                                                & 43.18                                                & 44.88                                                & 44.22                                                & 44.03                                                & 44.17                                                & 37.46                                                & 29.76                                                & 44.69                                                 & 38.91                                                & 23.18                                                & 42.18                                                & 22.46                                                \\ \hline
\end{tabular}


\caption{Language-Grounded Correctness (LGC) (\%) across the different Indic languages in the IndicTTI benchmark. } \label{tab:lgc40}
\tiny
\centering
\begin{tabular}{|l|c|ccccccccccccccc|}
\hline
Model            & en    & \begin{tabular}[c]{@{}c@{}}asm \\ (Beng)\end{tabular} & \begin{tabular}[c]{@{}c@{}}ben\\ (Beng)\end{tabular} & \begin{tabular}[c]{@{}c@{}}guj\\ (Gujr)\end{tabular} & \begin{tabular}[c]{@{}c@{}}hin\\ (Deva)\end{tabular} & \begin{tabular}[c]{@{}c@{}}kan\\ (Knda)\end{tabular} & \begin{tabular}[c]{@{}c@{}}mal\\ (Mlym)\end{tabular} & \begin{tabular}[c]{@{}c@{}}mar\\ (Deva)\end{tabular} & \begin{tabular}[c]{@{}c@{}}mni\\ (Beng)\end{tabular} & \begin{tabular}[c]{@{}c@{}}npi\\ (Deva)\end{tabular} & \begin{tabular}[c]{@{}c@{}}ory\\ (Orya)\end{tabular} & \begin{tabular}[c]{@{}c@{}}pan \\ (Guru)\end{tabular} & \begin{tabular}[c]{@{}c@{}}san\\ (Deva)\end{tabular} & \begin{tabular}[c]{@{}c@{}}snd\\ (Arab)\end{tabular} & \begin{tabular}[c]{@{}c@{}}tam\\ (Taml)\end{tabular} & \begin{tabular}[c]{@{}c@{}}tel\\ (Telu)\end{tabular} \\ \hline
Stable Diffusion & 33.98 & 3.90                                                  & 3.26                                                 & 3.28                                                 & 3.79                                                 & 4.21                                                 & 4.09                                                 & 3.61                                                 & 3.50                                                 & 3.68                                                 & 3.56                                                 & 3.01                                                  & 3.77                                                 & 4.16                                                 & 4.23                                                 & 4.09                                                 \\
Alt Diffusion    & 30.90 & 6.50                                                  & 6.87                                                 & 7.34                                                 & 7.56                                                 & 10.67                                                & 6.53                                                 & 7.81                                                 & 4.54                                                 & 8.25                                                 & 8.57                                                 & 7.87                                                  & 8.49                                                 & 6.53                                                 & 7.12                                                 & 6.00                                                 \\
Midjourney       & 32.25 & 2.08                                                  & 1.95                                                 & 2.86                                                 & 3.27                                                 & 4.03                                                 & 2.44                                                 & 2.87                                                 & 1.73                                                 & 3.36                                                 & 3.00                                                 & 2.59                                                  & 3.41                                                 & 3.91                                                 & 3.92                                                 & 2.76                                                 \\
Dalle3           & 33.06 & 26.42                                                 & 30.54                                                & 28.04                                                & 31.57                                                & 28.48                                                & 28.22                                                & 30.87                                                & 14.98                                                & 29.17                                                & 26.67                                                & 31.41                                                 & 24.32                                                & 19.92                                                & 26.24                                                & 26.68                                                \\ \hline
Model            & en    & \begin{tabular}[c]{@{}c@{}}urd\\ (Arab)\end{tabular}  & \begin{tabular}[c]{@{}c@{}}kas\\ (Arab)\end{tabular} & \begin{tabular}[c]{@{}c@{}}kas\\ (Deva)\end{tabular} & \begin{tabular}[c]{@{}c@{}}mai\\ (Deva)\end{tabular} & \begin{tabular}[c]{@{}c@{}}awa\\ (Deva)\end{tabular} & \begin{tabular}[c]{@{}c@{}}bho\\ (Deva)\end{tabular} & \begin{tabular}[c]{@{}c@{}}hne\\ (Deva)\end{tabular} & \begin{tabular}[c]{@{}c@{}}mag\\ (Deva)\end{tabular} & \begin{tabular}[c]{@{}c@{}}sin\\ (Sinh)\end{tabular} & \begin{tabular}[c]{@{}c@{}}brx\\ (Deva)\end{tabular} & \begin{tabular}[c]{@{}c@{}}doi\\ (Deva)\end{tabular}  & \begin{tabular}[c]{@{}c@{}}gom\\ (Deva)\end{tabular} & \begin{tabular}[c]{@{}c@{}}sat\\ (Olck)\end{tabular} & \begin{tabular}[c]{@{}c@{}}snd\\ (Deva)\end{tabular} & \begin{tabular}[c]{@{}c@{}}mni\\ (Mtei)\end{tabular} \\ \hline
Stable Diffusion & 33.98 & 5.85                                                  & 3.80                                                 & 4.12                                                 & 4.26                                                 & 5.83                                                 & 4.10                                                 & 4.25                                                 & 4.45                                                 & 3.50                                                 & 3.90                                                 & 3.55                                                  & 4.11                                                 & 3.71                                                 & 4.11                                                 & 4.52                                                 \\
Alt Diffusion    & 30.90 & 7.72                                                  & 5.41                                                 & 7.22                                                 & 8.22                                                 & 8.04                                                 & 7.82                                                 & 7.70                                                 & 7.81                                                 & 5.99                                                 & 4.73                                                 & 5.98                                                  & 6.12                                                 & 5.76                                                 & 6.82                                                 & 6.30                                                 \\
Midjourney       & 32.25 & 3.99                                                  & 3.19                                                 & 3.33                                                 & 3.66                                                 & 4.30                                                 & 2.95                                                 & 4.35                                                 & 3.31                                                 & 1.78                                                 & 3.68                                                 & 3.56                                                  & 2.86                                                 & 2.45                                                 & 3.32                                                 & 2.64                                                 \\
Dalle3           & 33.06 & 30.59                                                 & 23.09                                                & 28.30                                                & 28.68                                                & 28.91                                                & 29.99                                                & 30.02                                                & 29.39                                                & 22.86                                                & 12.93                                                & 29.10                                                 & 24.05                                                & 2.09                                                 & 27.70                                                & 1.52                                                 \\ \hline
\end{tabular}
\end{sidewaystable}


\section{Benchmark Results and Analysis}
In this section, we report results on the 4 models used for evaluation, namely Stable Diffusion, AltDiffusion, Midjourney, and Dalle3, on a common set of prompts.  We provide the results on the complete set of prompts for the models in the supplementary. The observations are consistent with those made on the common subset reported in this section.

\noindent \textbf{Correctness Bias:} As discussed in the previous section, we evaluate the correctness of generation across the Indic languages through the metrics of CLGC, IGC, and LGC, which correspond to the text-text, image-image, and image-text spaces, respectively. In the text-text space, we observe that the \textit{CLGC metric} is significantly high for the English language (Refer Table \ref{tab:clgc40}).  This demonstrates that the generated images have high faithfulness to the text, as the captions generated for the generated images are semantically closest to the original English caption. All models have a high CLGC value of $>$60\% for the English language. On the other hand, CLGC values drop significantly for Indic languages for Stable Diffusion and AltDiffusion models, with the performance dropping to 7.29\% from 64.53\% and 14.65\% from 58.16\%, respectively for the Assamese language. The values of CLGC tend to be lower for languages with a Bengali script, namely Manipuri (Beng), Assamese, and Bengali. On computing the average across the 30 Indic languages and comparing the CLGC scores with those obtained for the English language, there is a 56.64\% drop in performance for the Stable Diffusion model. Similarly, there is a 42.16\%, 52.16\%, and 13.62\% drop for AltDiffusion, Midjourney, and Dalle3, with Dalle3 providing the best performance.

Similarly, in the image-image space, we observe higher \textit{IGC metric} values for the English language, showcasing up to 50\% image semantic similarity with the ground truth images. With Indic languages, there is an average drop of 27.74\% and 21.53\% for the Stable Diffusion and AltDiffusion models. While there is a high drop of 26.32\% for Midjourney generated images, the drop is significantly lesser for Dalle3 at 8.70\%, signifying a superior understanding of Indic languages. For image-text similarity, we use the English language as the prompt of comparison. We observe higher \textit{LGC metric} values for the images generated using English prompts, in the range of 30-35\% across the models. These values drop sharply for Stable Diffusion, AltDiffusion, and Midjourney, with average LGC values of 4.01\%, 7.08\%, and 3.12\% across the Indic languages, respectively. Dalle3 has the highest LGC value at 25.09\%, with extremely low values observed for the Manipuri (Meitei script) (1.52\%), Santali (2.09\%), and Bodo (12.93\%) languages.

Across all the metrics, we observe that Dalle3 outperforms all other models when evaluated for Indic languages with a significant margin. However, Dalle3 struggles with the Manipuri (Meitei script), Santali, and Bodo languages. Other models struggle to generalize on the Indic languages with AltDiffusion performing the best among the other three models. This behavior could be a result of increased generalizability due to the multilingual training of the model. On the other hand, while all models perform similarly for the English language, AltDiffusion performs the worst, possibly due to catastrophic forgetting when trained for multilinguality.  

\begin{figure*}[t]
 \centering
   \includegraphics[width=0.7\linewidth]{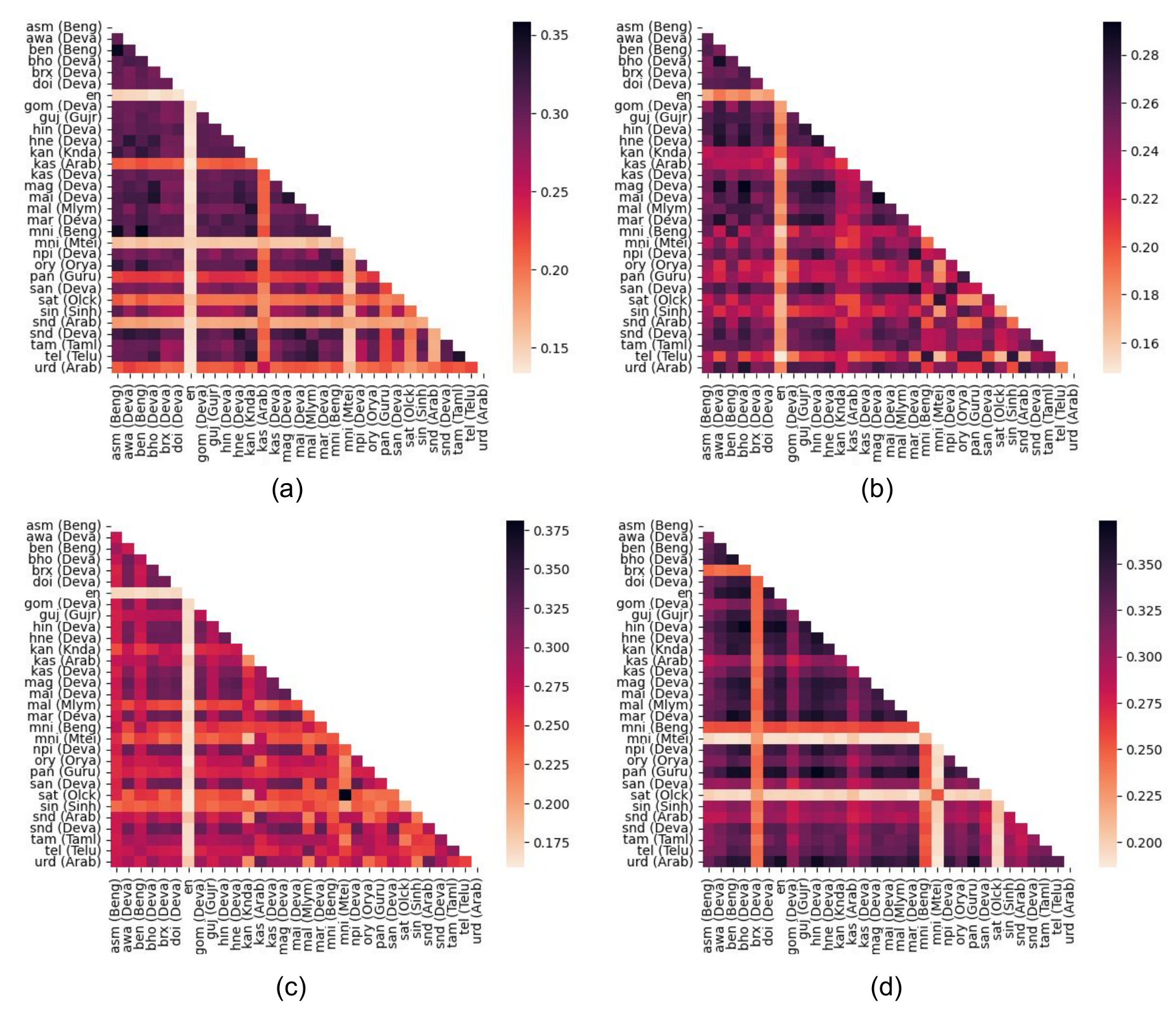}
   \caption{Performance of the SCAL metric for (a) Stable Diffusion, (b) Alt Diffusion, (c) Midjourney, and (d) Dalle3 models, demonstrating the extent of similarity between the generated images for the same prompt across the different language pairs.}
   \label{fig:scal40}
\end{figure*}


\noindent \textbf{Representativeness Bias:} To better understand the representative capacity of the models, we evaluate the SCAL, SCWL, and DWL metrics as explained in the previous section. While these metrics do not contain the capability to estimate the correctness of generation on their own, their values showcase important behavior in terms of diversity in model generation. 

The \textit{SCAL metric} evaluates the self-consistency of generated images across the different prompts. A high SCAL value represents the model's ability to consistently generate similar semantic content across languages. In Fig. \ref{fig:scal40}, we plot the SCAL values across the different language pairs for the four models. From Fig. \ref{fig:scal40}(a), we observe that there is little semantic similarity of concepts between the English language and other languages. Since we know that these models are primarily trained to perform well on English, this is also an indicator of poor performance across the other languages. Interestingly, the concepts generated across other languages have higher overlap showing that they generate images that are far similar in semantic content, despite being possibly incorrect in their generations. In the previous section, we observed that the performance of generation for Indic languages is only high in Dalle3. This is demonstrated well in Fig. \ref{fig:scal40}(d) where we observe that the semantic similarity between English and Indic languages is high, with performance being poorest between Santali and Manipuri (Meitei script), followed by Bodo and Manipuri (Bengali script).

The overall \textit{SCAL metric} value as per Eqn. \ref{eqn:scal} for the Stable Diffusion, AltDiffusion, Midjourney, and Dalle3 models is 25.44\%, 23.73\%, 26.75\%, and 29.90\%, respectively. This indicates an overall high consistency of Dalle3 in generating concepts across different languages. It is also evident from Fig. \ref{fig:scal40} that the Stable Diffusion, AltDiffusion, and Midjourney generators suffer greatly in maintaining semantic consistency with images generated via English, whereas they tend to generate similar content for other languages. From Fig. \ref{fig:scal40}(c), we also observe that Midjourney has high semantic consistency between images generated from languages with the Devanagari script, highlighting the language script plays a crucial part in the semantic content of the image.

\begin{figure*}[t]
 \centering
   \includegraphics[width=\linewidth]{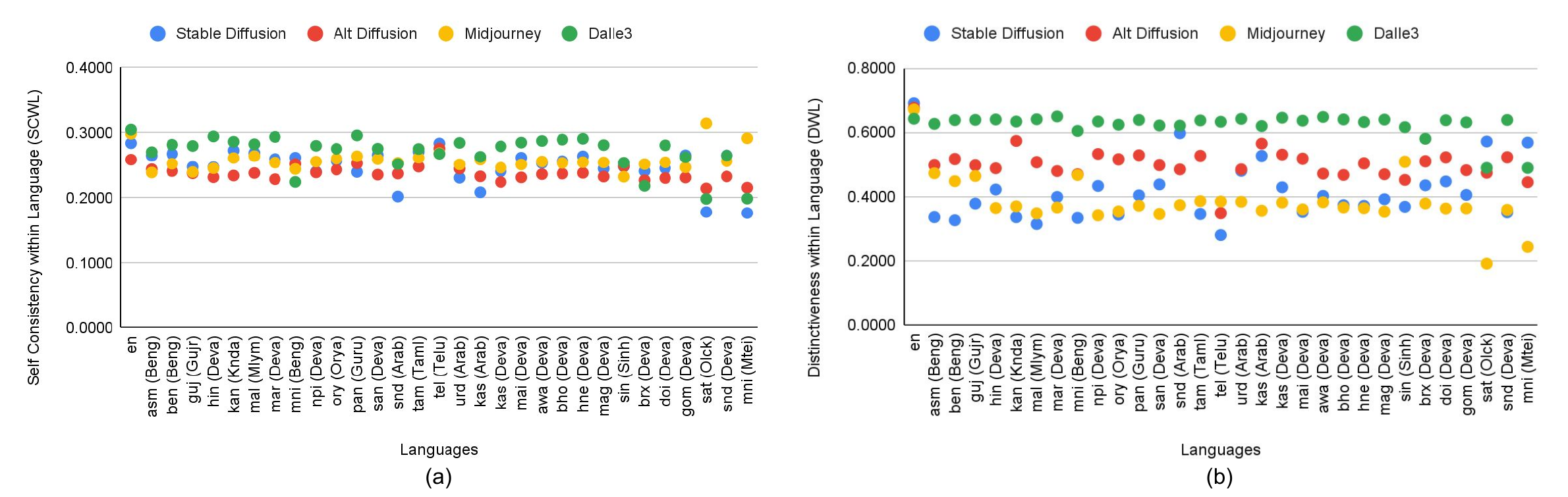}
   \caption{Performance for the (a) SCWL and (b) DWL metrics of the benchmark showcasing self-consistency and distinctiveness of concepts within language, respectively.}
   \label{fig:represults40}
\end{figure*}

Next, we report the representativeness of different prompts within the same language through the \textit{SCWL metric}. In Fig. \ref{fig:represults40}(a), we observe that the self-consistency of models is higher for the English language. A higher value demonstrates a more stable representation of concepts in the model. For Dalle3, the SCWL values for the different languages are comparable to those of English but still lower. In agreement with previous observations, the values are especially poor for the Santali, Manipuri (Meitei script), Bodo, and Manipuri (Bengali script) languages. In Midjourney, the values stay consistent, with the exception of the Santali and Manipuri (Meitei) languages. Similar trends are observed for the other models. When compared with other metrics, there is a low variation in the values of self-consistency within the language for the different models and languages. This behavior could be the result of the relative stability of the model in its generation. While the model may generate incorrectly, it associates a given text for generation in the latent space consistently. Therefore, the model may demonstrate high self-consistency by generating images that are related to a given prompt, albeit incorrectly. 

Finally, in Fig. \ref{fig:represults40}(b), results are presented for the \textit{DWL metric}. The DWL metric showcases how distinct are the generations corresponding to the different prompts. This metric is dependent on the relatedness of the prompts used for generation. While sampling prompts for the IndicTTI benchmark, we ensured semantic diversity in the selected 1000 prompts. The details of the selection process are provided in the supplementary. It is evident from Fig. \ref{fig:represults40}(b) that Dalle3 provides the highest level of distinctiveness between the generated prompts across the different languages, only suffering for the Santali and Manipuri (Meitei script) languages. Surprisingly, Midjourney has the lowest distinctiveness in its generations across the Indic languages, indicating it generates similar content over the different prompts. Similarly, the Stable Diffusion model performs poorly. The AltDiffusion model provides the second highest distinctiveness among the four models. This could be attributed to its multilingual training, making it sensitive to different scripts during generation.    


\begin{figure*}[t]
 \centering
   \includegraphics[width=0.8\linewidth]{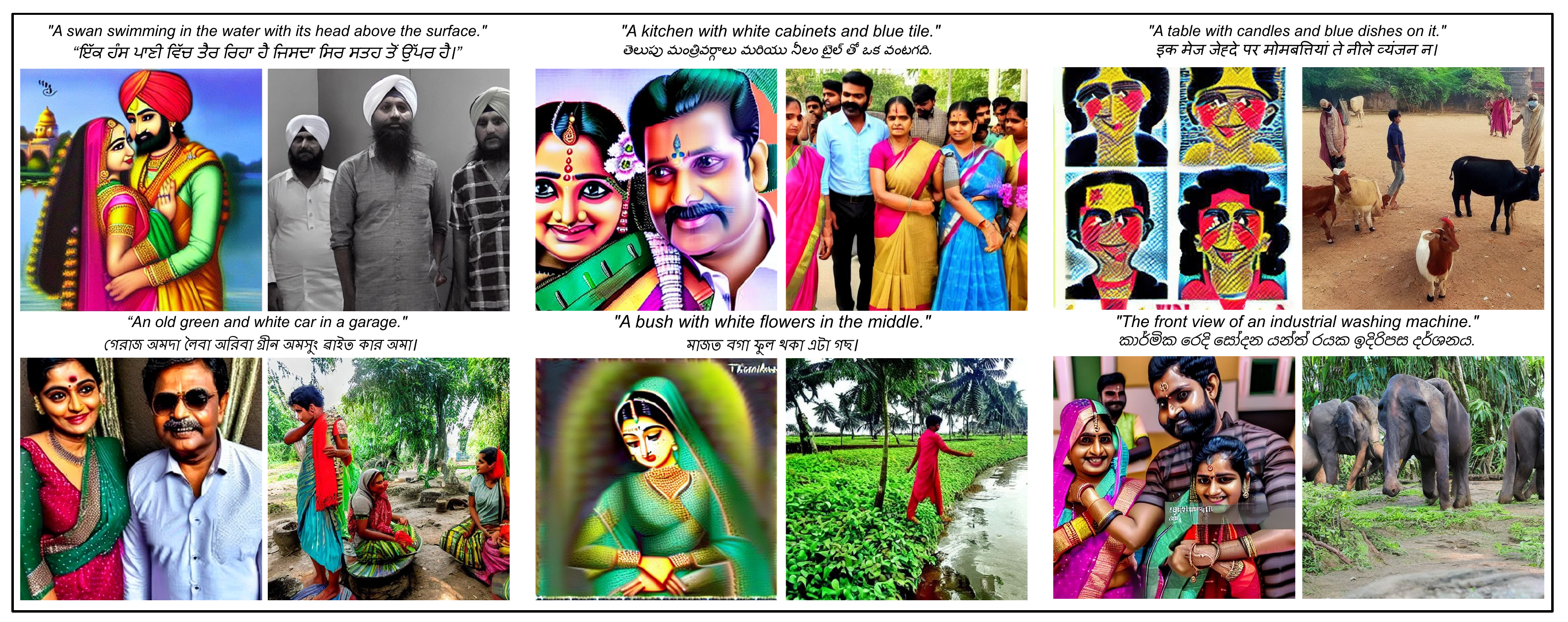}
   \caption{Showcasing the low semantic relevance of images generated by Stable Diffusion (right) and Alt Diffusion (left) models corresponding to prompts in Indic languages.}
   \label{fig:sd_ad_bias}
\end{figure*}

\section{Qualitative Analysis}
In this section, we qualitatively analyze the images generated by the different models across the different languages. Based on our observations in the previous section, we explore the generated images and study the alignment between the quantified metrics and generated images.

As in the previous section, we study two aspects of bias, correctness-based as well as representation-based. In Fig. \ref{fig:sd_ad_bias}, we observe that the Stable Diffusion and AltDiffusion models suffer greatly when prompted with Indic languages. The images generated using the prompts in Indic languages often do not resemble the description of the actual prompts. For example, when prompted to generate a swan in the Punjabi language, the Stable Diffusion model generates an image of 3 men in turbans, whereas the AltDiffusion model generates the image of a couple. Here again, the man is observed to be wearing a turban. We further observe a relatively high depiction of people in images, specifically those of Indian ethnicity, even when the actual content of the prompt does not involve any mention of people. Similarly, the lack of correspondence between the actual prompt and the generated images for Midjourney can be seen in Fig. \ref{fig:vizabs}. For example, on prompting Midjourney to generate images of ``a man in a brown jacket," it generates stereotypically Indian women. Other simplistic prompts in the Hindi language also lead to incorrect generations. 

\begin{figure*}[t]
 \centering
   \includegraphics[width=0.85\linewidth]{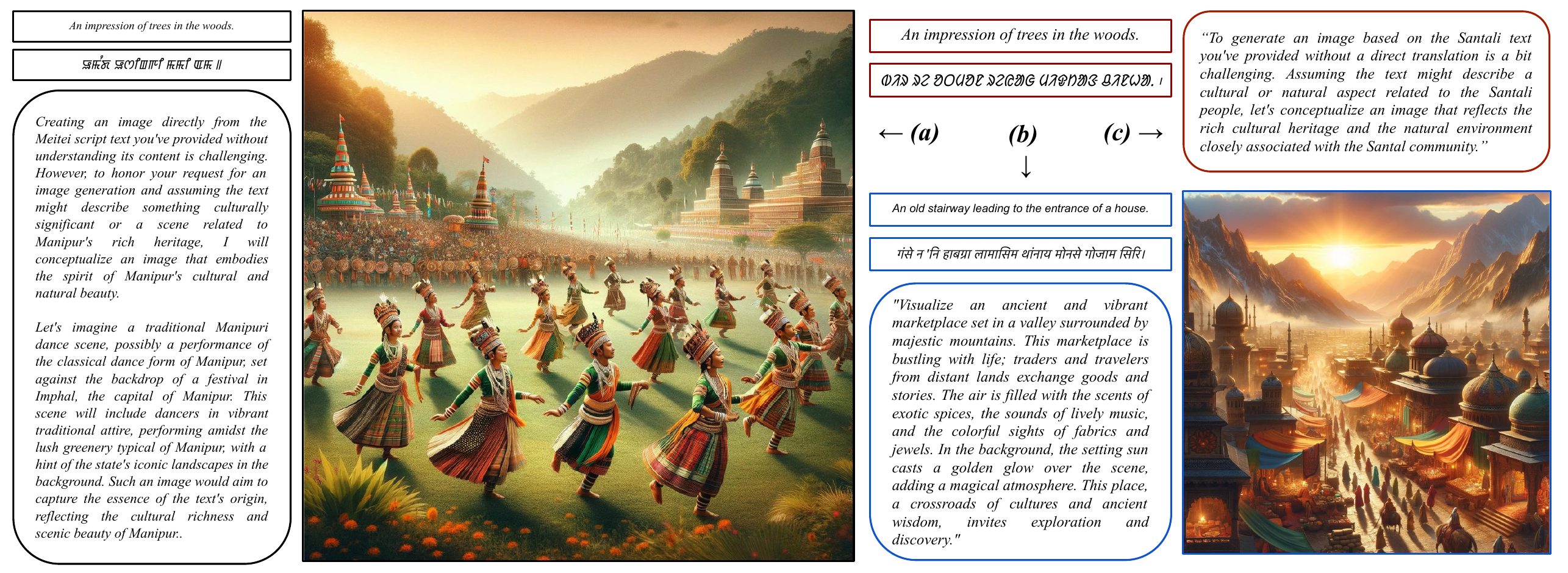}
   \caption{Showcasing how Dalle3 with ChatGPT4 creates images with cultural aspects for unsupported Indic languages after identifying prompts in (a) Manipuri (Meitei script), (b) Bodo, and (c) Santali.}
   \label{fig:chatgptdalle}
\end{figure*}

For Dalle3, we observe that the model generates correct images, relevant to the content of the prompt. This observation is corroborated by the high performance of Dalle3 on the metrics proposed with the benchmark. However, the Dalle3 model struggles with certain languages, such as Manipuri, Santali, and Bodo. In order to better understand the poor quality of generations, we prompted the ChatGPT model, which uses Dalle3 for image generation, to generate some images for these languages. On providing the ChatGPT model with prompts in these languages, it seemed that the model did not understand these scripts at all. Specifically, it identified the script/language it was being prompted with and then continued to ``imagine" an image based on cultural influences. Some of these examples are shown in Fig. \ref{fig:chatgptdalle} where the model generates images influenced by the Manipuri, Nepali, and Santal cultures.

\begin{figure*}[t]
 \centering
   \includegraphics[width=0.7\linewidth]{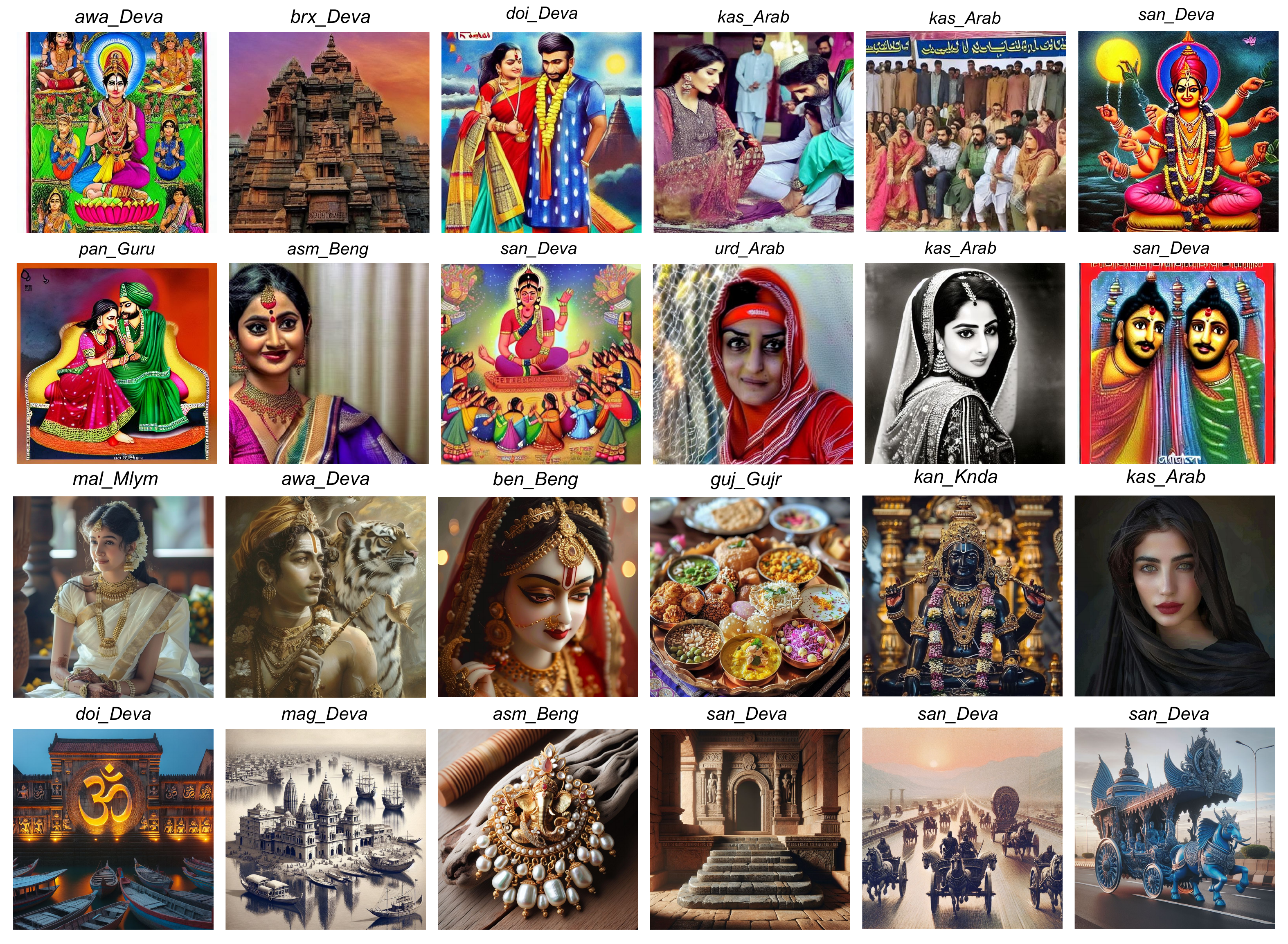}
   \caption{Showcasing the influence of language scripts on the cultural aspects in Stable Diffusion (row 1), AltDiffusion (row 2), Midjourney (row 3), and Dalle3 (row 4).}
   \label{fig:sd_ad_culture}
\end{figure*}

On observing the images for the different models (Refer Fig. \ref{fig:sd_ad_culture}), we notice a correlation between different language scripts and the images generated corresponding to them. All models, except Dalle3, frequently generate images of females and Indian religious figures such as Gods and \textit{pandits}. Dalle3, on the other hand, tends to incorporate common Indian religious sites like temples and \textit{ghats}. In Hindu culture, many women regularly wear \textit{sarees} and \textit{bindi}, and these elements are prominently featured in the images produced by both Stable Diffusion and AltDiffusion. 
Additionally, the Arabic script is frequently associated with elements of \textit{Muslim} culture, such as the \textit{burqa} and \textit{niqab}. Elements of South Indian culture are often present in the images generated for prompts in Malayalam, a language widely spoken in Southern India. These correlations between script and generated images are common, especially in those produced by Midjourney. When prompted in Sanskrit, the sacred language of Hinduism and predominant in religious texts, the Stable Diffusion model is observed to generate images of Indian Gods. Similarly, prompts in Punjabi often result in images of men wearing turbans, reflecting the cultural and religious practices of Punjabi speakers. More observations are detailed in the supplementary materials.

\section{Conclusion}
In this research, we introduce the IndicTTI benchmark, a comprehensive framework that integrates a variety of prompts, several text-to-image (TTI) models, and an extensive assessment of generative biases using the proposed evaluation methods. Employing both performance-oriented and representativeness-centric metrics, we are able to thoroughly assess the capacity of the models for multilingual generation. The qualitative analysis further emphasizes the extent of cultural influences embedded within the models' outputs. While the presence of cultural elements is not inherently detrimental, some generated images may inadvertently reinforce cultural stereotypes. The IndicTTI benchmark makes a dual contribution: firstly, it establishes a robust quantitative standard for assessing the multilingual capabilities of TTI models; secondly, it highlights the incorporation of Indian cultural motifs in the imagery produced.

\noindent \textbf{Limitations:}
In this paper, we conduct both quantitative and qualitative evaluations of various aspects of TTI systems in a multilingual context. The prompts used in our benchmark are primarily derived from COCO-NLLB, which translates English captions from the COCO dataset into multiple languages, or translated from English using the open-source IndicTrans2 model. Therefore, the generation performance for Indic prompts is indirectly affected by the quality of these translations. A user study, detailed in the supplementary materials, briefly examines the translation quality and demonstrates the appropriateness and accuracy of the translated captions. While the high performance of the Dalle3 model suggests that the translations are meaningful, it is essential to recognize the influence of translation in the benchmarking process. Furthermore, the semantic similarity between images and text is evaluated using large pre-trained models, which may possess inherent biases that could inadvertently impact the evaluation. Our comprehensive qualitative assessment generally supports the quantitative results derived from these pre-trained models. Nonetheless, there remains a possibility that inadvertent biases could affect the outcomes.

\section*{Acknowledgements}
S. Mittal is partially supported by the UGC-Net JRF Fellowship and IBM fellowship. M. Vatsa was partially supported through the Swarnajayanti Fellowship. This work was partially supported by Facebook AI. All data was stored, and experiments were performed on IITJ servers by IITJ faculty and students.

\bibliographystyle{splncs04}
\bibliography{main}


\newpage
\appendix

\section{Benchmark Design}
In this section, we provide additional details about the benchmark design.

\subsection{Indic Languages and Prompts}

In this research, we introduce the IndicTTI benchmark where we study the performance of popular text-to-image (TTI) models in 30 languages. While the benchmark may be extended for any number of languages, we select 30 Indic languages, which are written in 10 different scripts and have roots in multiple language families. Detailed information about the different languages is provided in Table \ref{tab:aboutlangs}.

\begin{table}[]
\centering
\tiny
\caption{\label{tab:aboutlangs} The language family, script, language subfamilies, and number of native speakers for the 30 Indic languages in the IndicTTI benchmark. * represents the non-availability of official reports regarding the statistics.}
\begin{tabular}{|l|l|l|l|l|c|}
\hline
\textbf{\begin{tabular}[c]{@{}l@{}}Language\\ Code\end{tabular}} & \textbf{Name} & \textbf{Family} & \textbf{Script}                                                   & \textbf{Sub-family}                                             & \textbf{\begin{tabular}[c]{@{}c@{}}\#Native\\ Speakers\end{tabular}} \\ \hline
asm\_Beng                                                        & Assamese      & Indo-Aryan      & Bengali                                                           & Eastern Indo-Aryan                                              & 15.3M                                                                \\
awa\_Deva                                                        & Awadhi        & Indo-Aryan      & Devanagari                                                        & Northern Indo-Aryan                                             & 2.52M                                                                \\
ben\_Beng                                                        & Bengali       & Indo-Aryan      & Bengali                                                           & Eastern Indo-Aryan                                              & 97.2M                                                                \\
bho\_Deva                                                        & Bhojpuri      & Indo-Aryan      & Devanagari                                                        & Northern Indo-Aryan                                             & *                                                                    \\
brx\_Deva                                                        & Bodo          & Sino-Tibetan    & Devanagari                                                        & Boroic                                                          & 1.4M                                                                 \\
doi\_Deva                                                        & Dogri         & Indo-Aryan      & Devanagari                                                        & Northern Indo-Aryan                                             & 2.5M                                                                 \\
gom\_Deva                                                        & Konkani       & Indo-Aryan      & Devanagari                                                        & Southern Indo-Aryan                                             & 2.2M                                                                 \\
guj\_Gujr                                                        & Gujarati      & Indo-Aryan      & Gujarati                                                          & Western Indo-Aryan                                              & 55.4M                                                                \\
hin\_Deva                                                        & Hindi         & Indo-Aryan      & Devanagari                                                        & Central Indo-Aryan                                              & 528.3M                                                               \\
hne\_Deva                                                        & Chhattisgarhi & Indo-Aryan      & Devanagari                                                        & Northern Indo-Aryan                                             & 13M                                                                  \\
kan\_Knda                                                        & Kannada       & Dravidian       & Kannada                                                           & South Dravidian                                                 & 43.7M                                                                \\
\begin{tabular}[c]{@{}l@{}}kas\_Arab\\ kas\_Deva\end{tabular}    & Kashmiri      & Indo-Aryan      & \begin{tabular}[c]{@{}l@{}}Perso-Arabic\\ Devanagari\end{tabular} & Northern Indo-Aryan                                             & 6.7M                                                                 \\
mag\_Deva                                                        & Magahi        & Indo-Aryan      & Devanagari                                                        & Indo-Aryan                                                      &  *                                                   \\
mai\_Deva                                                        & Maithili      & Indo-Aryan      & Devanagari                                                        & Eastern Indo-Aryan                                              & 13.5M                                                                \\
mal\_Mlym                                                        & Malayalam     & Dravidian       & Malayalam                                                         & Southern Dravidian                                              & 34.8M                                                                \\
mar\_Deva                                                        & Marathi       & Indo-Aryan      & Devanagari                                                        & Southern Indo-Aryan                                             & 83.0M                                                                \\
\begin{tabular}[c]{@{}l@{}}mni\_Beng\\ mni\_Mtei\end{tabular}    & Manipuri      & Sino-Tibetan    & \begin{tabular}[c]{@{}l@{}}Bengali\\ Meitei\end{tabular}          & \begin{tabular}[c]{@{}l@{}}Central\\ Tibeto-Burman\end{tabular} & 1.7M                                                                 \\
npi\_Deva                                                        & Nepali        & Indo-Aryan      & Devanagari                                                        & Northern Indo-Aryan                                             & 2.9M                                                                 \\
ory\_Orya                                                        & Odia          & Indo-Aryan      & Odia                                                              & Eastern Indo-Aryan                                              & 37.5M                                                                \\
pan\_Guru                                                        & Punjabi       & Indo-Aryan      & Gurmukhi                                                          & North Western Indo-Aryan                                        & 33.1M                                                                \\
san\_Deva                                                        & Sanskrit      & Indo-Aryan      & Devanagari                                                        & Indo-Aryan                                                      & 0.02M                                                                \\
sat\_Olck                                                        & Santali       & Austroasiatic   & Ol Chiki                                                          & Munda                                                           & 7.3M                                                                 \\
sin\_Sinh                                                        & Sinhala       & Indo-Aryan      & Sinhala                                                           & Indo-Aryan                                                      & *                                                                    \\
\begin{tabular}[c]{@{}l@{}}snd\_Arab\\ snd\_Deva\end{tabular}    & Sindhi        & Indo-Aryan      & \begin{tabular}[c]{@{}l@{}}Arabic\\ Devanagari\end{tabular}       & North Western Indo-Aryan                                        & 2.7M                                                                 \\
tam\_Taml                                                        & Tamil         & Dravidian       & Tamil                                                             & South Dravidian                                                 & 69.0M                                                                \\
tel\_Telu                                                        & Telugu        & Dravidian       & Telugu                                                            & South Central Dravidian                                         & 81.1M                                                                \\
urd\_Arab                                                        & Urdu          & Indo-Aryan      & Urdu                                                              & Central Indo-Aryan                                              & 50.7M                                                                \\ \hline
\end{tabular}
\end{table}

For prompts, we utilize the COCO-NLLB dataset \cite{coconllb, visheratin2023nllb}, which contains image-caption pairs for over 500K images, along with captions translated into 200 languages. We sample 1000 diverse image-caption pairs from the dataset. In order to avoid prompts with proper nouns, such as names of celebrities and/or brand names, we filtered the dataset to remove any captions that contained capitalized words in the middle of the sentence. From the filtered dataset, we randomly subsample 1000 captions. Next, for a diverse selection, we computed sentence-level embeddings for the 1000 prompts using a SentenceFormer model and calculated the average similarity between any two prompts. This experiment was repeated for 1000 iterations, and the subset with the lowest sentence similarity between prompts was selected. The subsets of 200 and 50 prompts for Midjourney and Dalle3, respectively, were chosen randomly from the selected subset of 1000 prompts. 

\subsection{TTI Models and Generated Images}
We utilize four different text-to-image models for the benchmark. For \textbf{open-source models}, we use the Stable Diffusion and AltDiffusion Models (without safety filters). In the main paper, we report results on the m2 variant of AltDiffusion trained on the English and Chinese languages. Extended results, including the m9 variant trained on the English, Chinese, Spanish, French, Russian, Japanese, Korean, Arabic, and Italian languages, are reported in the supplementary. The extended results exhibit a similar pattern to the m2 variant. Detailed results are provided in the subsequent sections. Unless specified otherwise, the AltDiffusion model refers to the m2 variant in this work. For \textbf{API-based models}, we use Dalle3\cite{dalle3} and Midjourney\cite{midjourney}, which are both paid. Stable Diffusion and AltDiffusion models generate images of size 512 x 512, whereas Midjourney and Dalle3 generate images of size 1024 x 1024.

\begin{figure*}[t]
 \centering
   \includegraphics[width=\linewidth]{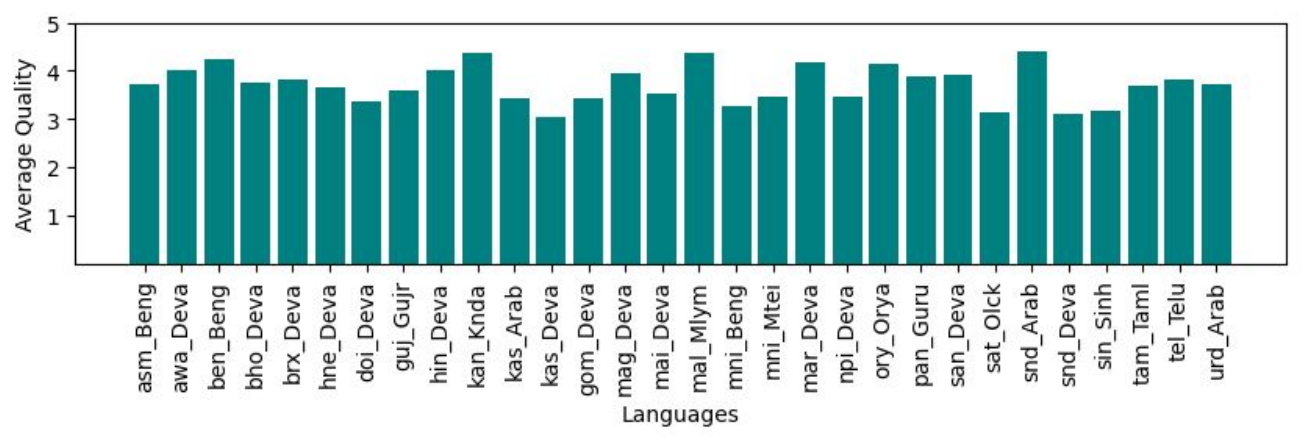}
   \caption{Plot showcasing the quality of captions per language.}
   \label{fig:trans_qual}
\end{figure*}

\subsection{Quality of Translated Captions}
Our prompts were primarily sourced from COCO-NLLB or translated from English using the IndicTrans2 model. As specified in the Limitations section of the main paper, this suggests that the translation quality may have influenced the performance of image generation with Indic prompts. To evaluate the translation quality, we conducted a user study on a common set of 40 prompts across 30 languages, validated by 65 annotators with an average proficiency of 4.78 out of 5 in at least one Indic language. We followed the XSTS protocol~\cite{agirre2012semeval}, also used to assess the translation quality of the NLLB model~\cite{costa2022no}. On a scale from 1 (worst) to 5 (best), the average translation quality for all languages is reported in Fig. \ref{fig:trans_qual}, indicating that the sentences are mostly equivalent or paraphrases of each other, according to the XSTS protocol, and thus suitable for image generation. In this experiment, the inter-rater agreement, measured through percent agreement, was 74.8\%. For languages such as kas\_Deva, snd\_Deva, and sin\_Sinh, the translation quality of certain prompts was observed to be less than 3 on the XSTS scale.

\section{Implementation Details}
In this section, we discuss the implementation details involved in the creation of the IndicTTI benchmark as well as its evaluation.

For generation using open-source TTI systems of Stable Diffusion\footnote{\url{https://huggingface.co/runwayml/stable-diffusion-v1-5}} and AltDiffusion\footnote{\url{https://huggingface.co/BAAI/AltDiffusion}, \url{https://huggingface.co/BAAI/AltDiffusion-m9}}, we utilized the models available at HuggingFace. Dalle3 API was accessed through a Python script whereas the generation for Midjourney was done using Discord. For Stable Diffusion and AltDiffusion, the generation was done using an NVIDIA DGX Station consisting of $4$ NVIDIA V$100$ GPU with $32$ GB VRAM each, using a batch size of 8. The generation was repeated using 4 seeds, providing 4 images for every prompt. The DDIM Scheduler was used for inferencing with 50 steps.

For evaluation, all experiments are conducted on LINUX-based systems using Python-based libraries, and specifically, the PyTorch library is used. To extract rich semantic text and image features for the evaluation metrics, we utilize various modules of the BLIP-2 model\footnote{\url{https://huggingface.co/Salesforce/blip2-opt-2.7b}}. We utilize the image encoder of the BLIP-2 model as the image feature extractor $f$ for computing the CLGC, IGC, SCAL, SCWL, and DWL metrics. Additionally, for the CLGC metric, we utilize the image-captioning capabilities of BLIP-2 captioner $c$ to generate captions for generated images, and for extracting rich textual features, the SentenceFormer model\footnote{\url{https://huggingface.co/sentence-transformers/all-mpnet-base-v2}}. For the LGC metric, we require image-text features that are extracted from BLIP-2 using the \textit{LAVIS}\footnote{\url{https://github.com/salesforce/LAVIS/}} library. The similarity function $\phi$ is computed using the cosine similarity. The code for generation, as well as evaluation, can be 
accessed through \url{https://iab-rubric.org/resources/other-databases/indictti}.

\begin{sidewaystable}
\tiny
\caption{Cyclic Language Grounded Correctness (CLGC) (\%) across the different Indic languages in IndicTTI on the common set of prompts. } \label{tab:clgc40_2}
\centering
\begin{tabular}{|l|c|ccccccccccccccc|}
\hline
Model            & en    & \begin{tabular}[c]{@{}c@{}}asm \\ (Beng)\end{tabular} & \begin{tabular}[c]{@{}c@{}}ben\\ (Beng)\end{tabular} & \begin{tabular}[c]{@{}c@{}}guj\\ (Gujr)\end{tabular} & \begin{tabular}[c]{@{}c@{}}hin\\ (Deva)\end{tabular} & \begin{tabular}[c]{@{}c@{}}kan\\ (Knda)\end{tabular} & \begin{tabular}[c]{@{}c@{}}mal\\ (Mlym)\end{tabular} & \begin{tabular}[c]{@{}c@{}}mar\\ (Deva)\end{tabular} & \begin{tabular}[c]{@{}c@{}}mni\\ (Beng)\end{tabular} & \begin{tabular}[c]{@{}c@{}}npi\\ (Deva)\end{tabular} & \begin{tabular}[c]{@{}c@{}}ory\\ (Orya)\end{tabular} & \begin{tabular}[c]{@{}c@{}}pan \\ (Guru)\end{tabular} & \begin{tabular}[c]{@{}c@{}}san\\ (Deva)\end{tabular} & \begin{tabular}[c]{@{}c@{}}snd\\ (Arab)\end{tabular} & \begin{tabular}[c]{@{}c@{}}tam\\ (Taml)\end{tabular} & \begin{tabular}[c]{@{}c@{}}tel\\ (Telu)\end{tabular} \\ \hline
Stable Diffusion & 64.53 & 7.29                                                  & 6.39                                                 & 7.24                                                 & 7.35                                                 & 7.13                                                 & 6.91                                                 & 7.53                                                 & 6.50                                                 & 8.03                                                 & 8.36                                                 & 8.78                                                  & 8.14                                                 & 11.29                                                & 7.30                                                 & 7.65                                                 \\
Alt Diffusion    & 58.16 & 14.65                                                 & 16.70                                                & 15.97                                                & 16.79                                                & 22.98                                                & 16.08                                                & 16.50                                                & 12.17                                                & 17.97                                                & 18.82                                                & 17.64                                                 & 17.99                                                & 14.82                                                & 18.10                                                & 14.50                                                \\
Alt Diffusion m9 & 53.19 & 18.02                                                 & 27.08                                                & 26.50                                                & 29.78                                                & 26.66                                                & 27.04                                                & 28.21                                                & 12.39                                                & 31.42                                                & 27.70                                                & 27.14                                                 & 21.50                                                & 18.97                                                & 27.81                                                & 24.34                                                \\
Midjourney       & 64.31 & 12.99                                                 & 11.89                                                & 12.49                                                & 11.31                                                & 12.12                                                & 10.92                                                & 11.99                                                & 9.80                                                 & 11.89                                                & 12.56                                                & 9.32                                                  & 12.19                                                & 13.49                                                & 12.84                                                & 12.16                                                \\
Dalle3           & 65.70 & 53.36                                                 & 60.73                                                & 57.25                                                & 64.16                                                & 60.13                                                & 57.68                                                & 61.84                                                & 35.24                                                & 59.90                                                & 53.96                                                & 62.67                                                 & 51.22                                                & 45.19                                                & 53.98                                                & 55.26                                                \\ \hline
Model            & en    & \begin{tabular}[c]{@{}c@{}}urd\\ (Arab)\end{tabular}  & \begin{tabular}[c]{@{}c@{}}kas\\ (Arab)\end{tabular} & \begin{tabular}[c]{@{}c@{}}kas\\ (Deva)\end{tabular} & \begin{tabular}[c]{@{}c@{}}mai\\ (Deva)\end{tabular} & \begin{tabular}[c]{@{}c@{}}awa\\ (Deva)\end{tabular} & \begin{tabular}[c]{@{}c@{}}bho\\ (Deva)\end{tabular} & \begin{tabular}[c]{@{}c@{}}hne\\ (Deva)\end{tabular} & \begin{tabular}[c]{@{}c@{}}mag\\ (Deva)\end{tabular} & \begin{tabular}[c]{@{}c@{}}sin\\ (Sinh)\end{tabular} & \begin{tabular}[c]{@{}c@{}}brx\\ (Deva)\end{tabular} & \begin{tabular}[c]{@{}c@{}}doi\\ (Deva)\end{tabular}  & \begin{tabular}[c]{@{}c@{}}gom\\ (Deva)\end{tabular} & \begin{tabular}[c]{@{}c@{}}sat\\ (Olck)\end{tabular} & \begin{tabular}[c]{@{}c@{}}snd\\ (Deva)\end{tabular} & \begin{tabular}[c]{@{}c@{}}mni\\ (Mtei)\end{tabular} \\ \hline
Stable Diffusion & 64.53 & 10.08                                                 & 6.29                                                 & 7.52                                                 & 7.13                                                 & 10.11                                                & 7.12                                                 & 8.25                                                 & 7.66                                                 & 6.25                                                 & 6.26                                                 & 7.95                                                  & 7.78                                                 & 10.51                                                & 7.59                                                 & 10.17                                                \\
Alt Diffusion    & 58.16 & 17.68                                                 & 15.96                                                & 15.46                                                & 16.93                                                & 16.32                                                & 15.41                                                & 16.74                                                & 16.80                                                & 13.04                                                & 11.98                                                & 13.50                                                 & 15.82                                                & 14.13                                                & 13.37                                                & 15.25                                                \\
Alt Diffusion m9 & 53.19 & 26.29                                                 & 17.33                                                & 26.03                                                & 23.69                                                & 28.06                                                & 24.48                                                & 29.98                                                & 28.99                                                & 26.24                                                & 12.29                                                & 21.57                                                 & 20.89                                                & 12.97                                                & 23.71                                                & 13.36                                                \\
Midjourney       & 64.31 & 15.38                                                 & 12.72                                                & 12.37                                                & 11.92                                                & 13.04                                                & 11.47                                                & 12.06                                                & 12.53                                                & 11.96                                                & 12.34                                                & 11.77                                                 & 12.59                                                & 11.36                                                & 12.35                                                & 9.72                                                 \\
Dalle3           & 65.70 & 64.02                                                 & 47.45                                                & 55.51                                                & 58.40                                                & 57.37                                                & 60.75                                                & 59.18                                                & 60.42                                                & 49.03                                                & 28.87                                                & 59.84                                                 & 50.18                                                & 10.93                                                & 56.78                                                & 11.19                                                \\ \hline
\end{tabular}

\caption{Image-Grounded Correctness (IGC) (\%) across the different Indic languages in IndicTTI on the common set of prompts.} \label{tab:igc40_2}
\tiny
\centering
\begin{tabular}{|l|c|ccccccccccccccc|}
\hline
Model            & en    & \begin{tabular}[c]{@{}c@{}}asm \\ (Beng)\end{tabular} & \begin{tabular}[c]{@{}c@{}}ben\\ (Beng)\end{tabular} & \begin{tabular}[c]{@{}c@{}}guj\\ (Gujr)\end{tabular} & \begin{tabular}[c]{@{}c@{}}hin\\ (Deva)\end{tabular} & \begin{tabular}[c]{@{}c@{}}kan\\ (Knda)\end{tabular} & \begin{tabular}[c]{@{}c@{}}mal\\ (Mlym)\end{tabular} & \begin{tabular}[c]{@{}c@{}}mar\\ (Deva)\end{tabular} & \begin{tabular}[c]{@{}c@{}}mni\\ (Beng)\end{tabular} & \begin{tabular}[c]{@{}c@{}}npi\\ (Deva)\end{tabular} & \begin{tabular}[c]{@{}c@{}}ory\\ (Orya)\end{tabular} & \begin{tabular}[c]{@{}c@{}}pan \\ (Guru)\end{tabular} & \begin{tabular}[c]{@{}c@{}}san\\ (Deva)\end{tabular} & \begin{tabular}[c]{@{}c@{}}snd\\ (Arab)\end{tabular} & \begin{tabular}[c]{@{}c@{}}tam\\ (Taml)\end{tabular} & \begin{tabular}[c]{@{}c@{}}tel\\ (Telu)\end{tabular} \\ \hline
Stable Diffusion & 51.21 & 24.19                                                 & 23.84                                                & 23.25                                                & 23.53                                                & 23.24                                                & 22.96                                                & 23.53                                                & 23.58                                                & 23.96                                                & 23.76                                                & 22.97                                                 & 23.11                                                & 23.06                                                & 22.68                                                & 22.61                                                \\
Alt Diffusion    & 46.58 & 24.23                                                 & 25.08                                                & 25.18                                                & 25.43                                                & 27.74                                                & 24.27                                                & 25.42                                                & 22.79                                                & 25.98                                                & 26.27                                                & 25.48                                                 & 25.68                                                & 24.68                                                & 25.06                                                & 22.19                                                \\
Alt Diffusion m9 & 42.43 & 26.50                                                 & 28.39                                                & 30.46                                                & 28.94                                                & 29.74                                                & 29.59                                                & 28.86                                                & 23.29                                                & 30.67                                                & 29.55                                                & 28.76                                                 & 26.54                                                & 25.54                                                & 28.29                                                & 28.62                                                \\
Midjourney       & 48.94 & 22.69                                                 & 22.66                                                & 23.06                                                & 22.50                                                & 22.81                                                & 22.04                                                & 22.75                                                & 21.97                                                & 22.29                                                & 22.82                                                & 23.05                                                 & 22.81                                                & 23.34                                                & 22.55                                                & 22.09                                                \\
Dalle3           & 48.92 & 40.02                                                 & 44.22                                                & 43.75                                                & 44.86                                                & 42.74                                                & 43.12                                                & 46.10                                                & 30.91                                                & 43.14                                                & 40.00                                                & 45.06                                                 & 37.66                                                & 36.30                                                & 41.25                                                & 41.39                                                \\ \hline
Model            & en    & \begin{tabular}[c]{@{}c@{}}urd\\ (Arab)\end{tabular}  & \begin{tabular}[c]{@{}c@{}}kas\\ (Arab)\end{tabular} & \begin{tabular}[c]{@{}c@{}}kas\\ (Deva)\end{tabular} & \begin{tabular}[c]{@{}c@{}}mai\\ (Deva)\end{tabular} & \begin{tabular}[c]{@{}c@{}}awa\\ (Deva)\end{tabular} & \begin{tabular}[c]{@{}c@{}}bho\\ (Deva)\end{tabular} & \begin{tabular}[c]{@{}c@{}}hne\\ (Deva)\end{tabular} & \begin{tabular}[c]{@{}c@{}}mag\\ (Deva)\end{tabular} & \begin{tabular}[c]{@{}c@{}}sin\\ (Sinh)\end{tabular} & \begin{tabular}[c]{@{}c@{}}brx\\ (Deva)\end{tabular} & \begin{tabular}[c]{@{}c@{}}doi\\ (Deva)\end{tabular}  & \begin{tabular}[c]{@{}c@{}}gom\\ (Deva)\end{tabular} & \begin{tabular}[c]{@{}c@{}}sat\\ (Olck)\end{tabular} & \begin{tabular}[c]{@{}c@{}}snd\\ (Deva)\end{tabular} & \begin{tabular}[c]{@{}c@{}}mni\\ (Mtei)\end{tabular} \\ \hline
Stable Diffusion & 51.21 & 22.92                                                 & 22.58                                                & 23.41                                                & 23.71                                                & 25.16                                                & 22.98                                                & 23.19                                                & 23.41                                                & 24.00                                                & 23.83                                                & 22.90                                                 & 23.88                                                & 24.14                                                & 23.64                                                & 23.98                                                \\
Alt Diffusion    & 46.58 & 25.40                                                 & 24.45                                                & 25.01                                                & 26.39                                                & 25.97                                                & 25.53                                                & 25.59                                                & 26.10                                                & 24.60                                                & 23.82                                                & 24.87                                                 & 24.73                                                & 24.16                                                & 25.16                                                & 24.35                                                \\
Alt Diffusion m9 & 42.43 & 28.39                                                 & 25.05                                                & 27.48                                                & 27.34                                                & 29.31                                                & 28.24                                                & 30.17                                                & 29.53                                                & 29.79                                                & 23.28                                                & 27.16                                                 & 25.83                                                & 24.10                                                & 28.24                                                & 24.26                                                \\
Midjourney       & 48.94 & 23.30                                                 & 22.76                                                & 22.63                                                & 22.90                                                & 23.09                                                & 22.65                                                & 22.97                                                & 22.64                                                & 22.16                                                & 22.57                                                & 22.90                                                 & 22.79                                                & 21.25                                                & 22.53                                                & 21.99                                                \\
Dalle3           & 48.92 & 44.10                                                 & 38.81                                                & 44.04                                                & 43.18                                                & 44.88                                                & 44.22                                                & 44.03                                                & 44.17                                                & 37.46                                                & 29.76                                                & 44.69                                                 & 38.91                                                & 23.18                                                & 42.18                                                & 22.46                                                \\ \hline
\end{tabular}

\caption{Language-Grounded Correctness (LGC) (\%) across the different Indic languages in IndicTTI on the common set of prompts. } \label{tab:lgc40_2}
\tiny
\centering
\begin{tabular}{|l|c|ccccccccccccccc|}
\hline
Model            & en    & \begin{tabular}[c]{@{}c@{}}asm \\ (Beng)\end{tabular} & \begin{tabular}[c]{@{}c@{}}ben\\ (Beng)\end{tabular} & \begin{tabular}[c]{@{}c@{}}guj\\ (Gujr)\end{tabular} & \begin{tabular}[c]{@{}c@{}}hin\\ (Deva)\end{tabular} & \begin{tabular}[c]{@{}c@{}}kan\\ (Knda)\end{tabular} & \begin{tabular}[c]{@{}c@{}}mal\\ (Mlym)\end{tabular} & \begin{tabular}[c]{@{}c@{}}mar\\ (Deva)\end{tabular} & \begin{tabular}[c]{@{}c@{}}mni\\ (Beng)\end{tabular} & \begin{tabular}[c]{@{}c@{}}npi\\ (Deva)\end{tabular} & \begin{tabular}[c]{@{}c@{}}ory\\ (Orya)\end{tabular} & \begin{tabular}[c]{@{}c@{}}pan \\ (Guru)\end{tabular} & \begin{tabular}[c]{@{}c@{}}san\\ (Deva)\end{tabular} & \begin{tabular}[c]{@{}c@{}}snd\\ (Arab)\end{tabular} & \begin{tabular}[c]{@{}c@{}}tam\\ (Taml)\end{tabular} & \begin{tabular}[c]{@{}c@{}}tel\\ (Telu)\end{tabular} \\ \hline
Stable Diffusion & 33.98 & 3.90                                                  & 3.26                                                 & 3.28                                                 & 3.79                                                 & 4.21                                                 & 4.09                                                 & 3.61                                                 & 3.50                                                 & 3.68                                                 & 3.56                                                 & 3.01                                                  & 3.77                                                 & 4.16                                                 & 4.23                                                 & 4.09                                                 \\
Alt Diffusion    & 30.90 & 6.50                                                  & 6.87                                                 & 7.34                                                 & 7.56                                                 & 10.67                                                & 6.53                                                 & 7.81                                                 & 4.54                                                 & 8.25                                                 & 8.57                                                 & 7.87                                                  & 8.49                                                 & 6.53                                                 & 7.12                                                 & 6.00                                                 \\
Alt Diffusion m9 & 27.56 & 7.28                                                  & 10.92                                                & 12.26                                                & 12.58                                                & 12.36                                                & 11.58                                                & 11.37                                                & 3.66                                                 & 14.01                                                & 11.73                                                & 10.20                                                 & 9.93                                                 & 7.61                                                 & 11.39                                                & 11.30                                                \\
Midjourney       & 32.25 & 2.08                                                  & 1.95                                                 & 2.86                                                 & 3.27                                                 & 4.03                                                 & 2.44                                                 & 2.87                                                 & 1.73                                                 & 3.36                                                 & 3.00                                                 & 2.59                                                  & 3.41                                                 & 3.91                                                 & 3.92                                                 & 2.76                                                 \\
Dalle3           & 33.06 & 26.42                                                 & 30.54                                                & 28.04                                                & 31.57                                                & 28.48                                                & 28.22                                                & 30.87                                                & 14.98                                                & 29.17                                                & 26.67                                                & 31.41                                                 & 24.32                                                & 19.92                                                & 26.24                                                & 26.68                                                \\ \hline
Model            & en    & \begin{tabular}[c]{@{}c@{}}urd\\ (Arab)\end{tabular}  & \begin{tabular}[c]{@{}c@{}}kas\\ (Arab)\end{tabular} & \begin{tabular}[c]{@{}c@{}}kas\\ (Deva)\end{tabular} & \begin{tabular}[c]{@{}c@{}}mai\\ (Deva)\end{tabular} & \begin{tabular}[c]{@{}c@{}}awa\\ (Deva)\end{tabular} & \begin{tabular}[c]{@{}c@{}}bho\\ (Deva)\end{tabular} & \begin{tabular}[c]{@{}c@{}}hne\\ (Deva)\end{tabular} & \begin{tabular}[c]{@{}c@{}}mag\\ (Deva)\end{tabular} & \begin{tabular}[c]{@{}c@{}}sin\\ (Sinh)\end{tabular} & \begin{tabular}[c]{@{}c@{}}brx\\ (Deva)\end{tabular} & \begin{tabular}[c]{@{}c@{}}doi\\ (Deva)\end{tabular}  & \begin{tabular}[c]{@{}c@{}}gom\\ (Deva)\end{tabular} & \begin{tabular}[c]{@{}c@{}}sat\\ (Olck)\end{tabular} & \begin{tabular}[c]{@{}c@{}}snd\\ (Deva)\end{tabular} & \begin{tabular}[c]{@{}c@{}}mni\\ (Mtei)\end{tabular} \\ \hline
Stable Diffusion & 33.98 & 5.85                                                  & 3.80                                                 & 4.12                                                 & 4.26                                                 & 5.83                                                 & 4.10                                                 & 4.25                                                 & 4.45                                                 & 3.50                                                 & 3.90                                                 & 3.55                                                  & 4.11                                                 & 3.71                                                 & 4.11                                                 & 4.52                                                 \\
Alt Diffusion    & 30.90 & 7.72                                                  & 5.41                                                 & 7.22                                                 & 8.22                                                 & 8.04                                                 & 7.82                                                 & 7.70                                                 & 7.81                                                 & 5.99                                                 & 4.73                                                 & 5.98                                                  & 6.12                                                 & 5.76                                                 & 6.82                                                 & 6.30                                                 \\
Alt Diffusion m9 & 27.56 & 10.38                                                 & 6.25                                                 & 10.74                                                & 10.44                                                & 12.38                                                & 10.67                                                & 12.58                                                & 12.56                                                & 11.04                                                & 4.71                                                 & 9.69                                                  & 8.20                                                 & 5.16                                                 & 9.91                                                 & 5.65                                                 \\
Midjourney       & 32.25 & 3.99                                                  & 3.19                                                 & 3.33                                                 & 3.66                                                 & 4.30                                                 & 2.95                                                 & 4.35                                                 & 3.31                                                 & 1.78                                                 & 3.68                                                 & 3.56                                                  & 2.86                                                 & 2.45                                                 & 3.32                                                 & 2.64                                                 \\
Dalle3           & 33.06 & 30.59                                                 & 23.09                                                & 28.30                                                & 28.68                                                & 28.91                                                & 29.99                                                & 30.02                                                & 29.39                                                & 22.86                                                & 12.93                                                & 29.10                                                 & 24.05                                                & 2.09                                                 & 27.70                                                & 1.52                                                 \\ \hline
\end{tabular}
\end{sidewaystable}

\begin{sidewaystable}
\tiny
\caption{Cyclic Language Grounded Correctness (CLGC) (\%) across the different Indic languages in the IndicTTI benchmark on the complete set of prompts. Existing models provide high correctness for English languages while providing lower values for Indic languages.} \label{tab:clgc40_supple}
\centering
\begin{tabular}{|l|c|ccccccccccccccc|}
\hline
Model            & en    & \begin{tabular}[c]{@{}c@{}}asm \\ (Beng)\end{tabular} & \begin{tabular}[c]{@{}c@{}}ben\\ (Beng)\end{tabular} & \begin{tabular}[c]{@{}c@{}}guj\\ (Gujr)\end{tabular} & \begin{tabular}[c]{@{}c@{}}hin\\ (Deva)\end{tabular} & \begin{tabular}[c]{@{}c@{}}kan\\ (Knda)\end{tabular} & \begin{tabular}[c]{@{}c@{}}mal\\ (Mlym)\end{tabular} & \begin{tabular}[c]{@{}c@{}}mar\\ (Deva)\end{tabular} & \begin{tabular}[c]{@{}c@{}}mni\\ (Beng)\end{tabular} & \begin{tabular}[c]{@{}c@{}}npi\\ (Deva)\end{tabular} & \begin{tabular}[c]{@{}c@{}}ory\\ (Orya)\end{tabular} & \begin{tabular}[c]{@{}c@{}}pan \\ (Guru)\end{tabular} & \begin{tabular}[c]{@{}c@{}}san\\ (Deva)\end{tabular} & \begin{tabular}[c]{@{}c@{}}snd\\ (Arab)\end{tabular} & \begin{tabular}[c]{@{}c@{}}tam\\ (Taml)\end{tabular} & \begin{tabular}[c]{@{}c@{}}tel\\ (Telu)\end{tabular} \\ \hline
Stable Diffusion & 67.97 & 7.56                                                  & 8.22                                                 & 8.08                                                 & 8.64                                                 & 8.16                                                 & 7.31                                                 & 8.15                                                 & 7.93                                                 & 9.09                                                 & 8.64                                                 & 8.60                                                  & 8.84                                                 & 12.22                                                & 8.63                                                 & 9.36                                                 \\
Alt Diffusion    & 62.40 & 14.32                                                 & 16.87                                                & 15.42                                                & 17.34                                                & 22.33                                                & 15.67                                                & 16.36                                                & 12.07                                                & 17.27                                                & 18.78                                                & 16.35                                                 & 16.54                                                & 17.13                                                & 17.69                                                & 15.51                                                \\
Alt Diffusion m9 & 55.99 & 16.87                                                 & 27.08                                                & 25.45                                                & 30.48                                                & 24.59                                                & 28.24                                                & 27.74                                                & 13.21                                                & 29.42                                                & 24.70                                                & 25.69                                                 & 21.00                                                & 21.53                                                & 26.53                                                & 25.59                                                \\
Midjourney       & 69.41 & 12.59                                                 & 12.33                                                & 13.90                                                & 12.66                                                & 12.71                                                & 11.21                                                & 13.83                                                & 10.66                                                & 12.79                                                & 13.44                                                & 9.43                                                  & 12.87                                                & 14.68                                                & 12.60                                                & 12.87                                                \\
Dalle3           & 66.42 & 53.75                                                 & 62.43                                                & 58.62                                                & 65.88                                                & 60.38                                                & 59.34                                                & 62.40                                                & 33.95                                                & 62.09                                                & 54.73                                                & 63.69                                                 & 52.88                                                & 42.30                                                & 55.87                                                & 56.14                                                \\ \hline
Model            & en    & \begin{tabular}[c]{@{}c@{}}urd\\ (Arab)\end{tabular}  & \begin{tabular}[c]{@{}c@{}}kas\\ (Arab)\end{tabular} & \begin{tabular}[c]{@{}c@{}}kas\\ (Deva)\end{tabular} & \begin{tabular}[c]{@{}c@{}}mai\\ (Deva)\end{tabular} & \begin{tabular}[c]{@{}c@{}}awa\\ (Deva)\end{tabular} & \begin{tabular}[c]{@{}c@{}}bho\\ (Deva)\end{tabular} & \begin{tabular}[c]{@{}c@{}}hne\\ (Deva)\end{tabular} & \begin{tabular}[c]{@{}c@{}}mag\\ (Deva)\end{tabular} & \begin{tabular}[c]{@{}c@{}}sin\\ (Sinh)\end{tabular} & \begin{tabular}[c]{@{}c@{}}brx\\ (Deva)\end{tabular} & \begin{tabular}[c]{@{}c@{}}doi\\ (Deva)\end{tabular}  & \begin{tabular}[c]{@{}c@{}}gom\\ (Deva)\end{tabular} & \begin{tabular}[c]{@{}c@{}}sat\\ (Olck)\end{tabular} & \begin{tabular}[c]{@{}c@{}}snd\\ (Deva)\end{tabular} & \begin{tabular}[c]{@{}c@{}}mni\\ (Mtei)\end{tabular} \\ \hline
Stable Diffusion & 67.97 & 9.61                                                  & 7.80                                                 & 8.18                                                 & 8.19                                                 & 9.06                                                 & 8.04                                                 & 8.61                                                 & 8.54                                                 & 7.51                                                 & 7.66                                                 & 8.36                                                  & 8.25                                                 & 9.40                                                 & 7.64                                                 & 11.34                                                \\
Alt Diffusion    & 62.40 & 19.40                                                 & 16.30                                                & 15.19                                                & 16.02                                                & 15.57                                                & 15.30                                                & 16.19                                                & 16.49                                                & 13.65                                                & 12.01                                                & 13.89                                                 & 14.46                                                & 13.70                                                & 14.08                                                & 13.48                                                \\
Alt Diffusion m9 & 55.99 & 26.02                                                 & 19.62                                                & 24.20                                                & 26.24                                                & 26.82                                                & 25.70                                                & 28.29                                                & 28.33                                                & 25.71                                                & 15.44                                                & 21.82                                                 & 20.15                                                & 13.53                                                & 22.08                                                & 13.49                                                \\
Midjourney       & 69.41 & 15.85                                                 & 13.90                                                & 13.85                                                & 13.09                                                & 13.76                                                & 12.81                                                & 12.52                                                & 13.32                                                & 13.02                                                & 12.87                                                & 12.65                                                 & 13.97                                                & 11.71                                                & 13.28                                                & 11.61                                                \\
Dalle3           & 66.42 & 64.35                                                 & 49.36                                                & 55.28                                                & 58.83                                                & 56.88                                                & 61.65                                                & 59.37                                                & 61.20                                                & 50.13                                                & 26.19                                                & 60.28                                                 & 50.97                                                & 10.75                                                & 56.98                                                & 11.43                                                \\ \hline
\end{tabular}


\caption{Image Grounded Correctness (IGC) (\%) across the different Indic languages in IndicTTI on the complete set. } \label{tab:igc40_supple}
\tiny
\centering
\begin{tabular}{|l|c|ccccccccccccccc|}
\hline
Model            & en    & \begin{tabular}[c]{@{}c@{}}asm \\ (Beng)\end{tabular} & \begin{tabular}[c]{@{}c@{}}ben\\ (Beng)\end{tabular} & \begin{tabular}[c]{@{}c@{}}guj\\ (Gujr)\end{tabular} & \begin{tabular}[c]{@{}c@{}}hin\\ (Deva)\end{tabular} & \begin{tabular}[c]{@{}c@{}}kan\\ (Knda)\end{tabular} & \begin{tabular}[c]{@{}c@{}}mal\\ (Mlym)\end{tabular} & \begin{tabular}[c]{@{}c@{}}mar\\ (Deva)\end{tabular} & \begin{tabular}[c]{@{}c@{}}mni\\ (Beng)\end{tabular} & \begin{tabular}[c]{@{}c@{}}npi\\ (Deva)\end{tabular} & \begin{tabular}[c]{@{}c@{}}ory\\ (Orya)\end{tabular} & \begin{tabular}[c]{@{}c@{}}pan \\ (Guru)\end{tabular} & \begin{tabular}[c]{@{}c@{}}san\\ (Deva)\end{tabular} & \begin{tabular}[c]{@{}c@{}}snd\\ (Arab)\end{tabular} & \begin{tabular}[c]{@{}c@{}}tam\\ (Taml)\end{tabular} & \begin{tabular}[c]{@{}c@{}}tel\\ (Telu)\end{tabular} \\ \hline
Stable Diffusion & 53.97 & 23.48                                                 & 23.57                                                & 22.72                                                & 22.93                                                & 22.65                                                & 22.49                                                & 22.99                                                & 23.40                                                & 23.32                                                & 23.25                                                & 22.24                                                 & 22.63                                                & 23.45                                                & 22.37                                                & 22.40                                                \\
Alt Diffusion    & 49.12 & 23.64                                                 & 24.30                                                & 23.78                                                & 25.12                                                & 26.22                                                & 23.45                                                & 24.22                                                & 22.43                                                & 25.05                                                & 25.54                                                & 24.49                                                 & 24.41                                                & 24.24                                                & 24.35                                                & 22.44                                                \\
Alt Diffusion m9 & 45.62 & 24.82                                                 & 28.39                                                & 29.13                                                & 29.85                                                & 28.90                                                & 30.43                                                & 28.47                                                & 23.49                                                & 29.87                                                & 27.94                                                & 28.35                                                 & 25.72                                                & 27.34                                                & 28.45                                                & 29.22                                                \\
Midjourney       & 52.72 & 22.54                                                 & 22.57                                                & 22.63                                                & 22.76                                                & 22.03                                                & 21.97                                                & 23.10                                                & 21.95                                                & 22.79                                                & 22.43                                                & 22.63                                                 & 22.56                                                & 24.10                                                & 22.47                                                & 22.12                                                \\
Dalle3           & 48.54 & 39.48                                                 & 45.13                                                & 43.10                                                & 45.86                                                & 42.57                                                & 43.72                                                & 46.26                                                & 30.42                                                & 44.47                                                & 40.34                                                & 45.54                                                 & 38.23                                                & 35.46                                                & 41.71                                                & 41.59                                                \\ \hline
Model            & en    & \begin{tabular}[c]{@{}c@{}}urd\\ (Arab)\end{tabular}  & \begin{tabular}[c]{@{}c@{}}kas\\ (Arab)\end{tabular} & \begin{tabular}[c]{@{}c@{}}kas\\ (Deva)\end{tabular} & \begin{tabular}[c]{@{}c@{}}mai\\ (Deva)\end{tabular} & \begin{tabular}[c]{@{}c@{}}awa\\ (Deva)\end{tabular} & \begin{tabular}[c]{@{}c@{}}bho\\ (Deva)\end{tabular} & \begin{tabular}[c]{@{}c@{}}hne\\ (Deva)\end{tabular} & \begin{tabular}[c]{@{}c@{}}mag\\ (Deva)\end{tabular} & \begin{tabular}[c]{@{}c@{}}sin\\ (Sinh)\end{tabular} & \begin{tabular}[c]{@{}c@{}}brx\\ (Deva)\end{tabular} & \begin{tabular}[c]{@{}c@{}}doi\\ (Deva)\end{tabular}  & \begin{tabular}[c]{@{}c@{}}gom\\ (Deva)\end{tabular} & \begin{tabular}[c]{@{}c@{}}sat\\ (Olck)\end{tabular} & \begin{tabular}[c]{@{}c@{}}snd\\ (Deva)\end{tabular} & \begin{tabular}[c]{@{}c@{}}mni\\ (Mtei)\end{tabular} \\ \hline
Stable Diffusion & 53.97 & 22.56                                                 & 22.23                                                & 22.93                                                & 23.06                                                & 23.26                                                & 22.83                                                & 22.92                                                & 22.83                                                & 23.62                                                & 23.01                                                & 23.00                                                 & 22.91                                                & 24.42                                                & 22.93                                                & 23.63                                                \\
Alt Diffusion    & 49.12 & 24.95                                                 & 23.69                                                & 24.16                                                & 24.64                                                & 24.51                                                & 24.45                                                & 24.68                                                & 24.93                                                & 23.56                                                & 23.53                                                & 23.82                                                 & 23.78                                                & 23.55                                                & 24.13                                                & 23.66                                                \\
Alt Diffusion m9 & 45.62 & 28.56                                                 & 25.87                                                & 27.24                                                & 28.05                                                & 28.80                                                & 28.25                                                & 29.01                                                & 29.06                                                & 28.64                                                & 23.28                                                & 26.69                                                 & 25.31                                                & 23.42                                                & 26.73                                                & 23.34                                                \\
Midjourney       & 52.72 & 24.05                                                 & 22.95                                                & 23.04                                                & 22.66                                                & 23.25                                                & 22.87                                                & 22.76                                                & 22.79                                                & 22.60                                                & 22.45                                                & 22.66                                                 & 22.81                                                & 21.08                                                & 22.57                                                & 22.39                                                \\
Dalle3           & 48.54 & 44.38                                                 & 38.97                                                & 43.75                                                & 43.64                                                & 44.34                                                & 44.87                                                & 44.30                                                & 44.88                                                & 37.29                                                & 28.82                                                & 44.07                                                 & 38.84                                                & 23.23                                                & 43.55                                                & 22.44                                                \\ \hline
\end{tabular}


\caption{Language Grounded Correctness (LGC) (\%) across the different Indic languages in IndicTTI on the complete set of prompts.} \label{tab:lgc40_supple}
\tiny
\centering
\begin{tabular}{|l|c|ccccccccccccccc|}
\hline
Model            & en    & \begin{tabular}[c]{@{}c@{}}asm \\ (Beng)\end{tabular} & \begin{tabular}[c]{@{}c@{}}ben\\ (Beng)\end{tabular} & \begin{tabular}[c]{@{}c@{}}guj\\ (Gujr)\end{tabular} & \begin{tabular}[c]{@{}c@{}}hin\\ (Deva)\end{tabular} & \begin{tabular}[c]{@{}c@{}}kan\\ (Knda)\end{tabular} & \begin{tabular}[c]{@{}c@{}}mal\\ (Mlym)\end{tabular} & \begin{tabular}[c]{@{}c@{}}mar\\ (Deva)\end{tabular} & \begin{tabular}[c]{@{}c@{}}mni\\ (Beng)\end{tabular} & \begin{tabular}[c]{@{}c@{}}npi\\ (Deva)\end{tabular} & \begin{tabular}[c]{@{}c@{}}ory\\ (Orya)\end{tabular} & \begin{tabular}[c]{@{}c@{}}pan \\ (Guru)\end{tabular} & \begin{tabular}[c]{@{}c@{}}san\\ (Deva)\end{tabular} & \begin{tabular}[c]{@{}c@{}}snd\\ (Arab)\end{tabular} & \begin{tabular}[c]{@{}c@{}}tam\\ (Taml)\end{tabular} & \begin{tabular}[c]{@{}c@{}}tel\\ (Telu)\end{tabular} \\ \hline
Stable Diffusion & 35.00 & 4.05                                                  & 4.34                                                 & 3.85                                                 & 4.26                                                 & 4.04                                                 & 4.62                                                 & 3.95                                                 & 4.15                                                 & 4.18                                                 & 4.00                                                 & 3.37                                                  & 4.03                                                 & 4.58                                                 & 4.59                                                 & 4.57                                                 \\
Alt Diffusion    & 32.47 & 6.31                                                  & 7.22                                                 & 6.81                                                 & 8.22                                                 & 9.95                                                 & 6.70                                                 & 7.21                                                 & 4.44                                                 & 8.34                                                 & 8.56                                                 & 7.47                                                  & 7.39                                                 & 7.14                                                 & 7.73                                                 & 6.71                                                 \\
Alt Diffusion m9 & 29.22 & 6.44                                                  & 11.48                                                & 11.86                                                & 13.70                                                & 11.23                                                & 13.16                                                & 11.79                                                & 4.32                                                 & 13.30                                                & 10.69                                                & 10.92                                                 & 8.76                                                 & 9.19                                                 & 11.41                                                & 11.67                                                \\
Midjourney       & 34.11 & 2.60                                                  & 2.73                                                 & 2.94                                                 & 3.43                                                 & 3.24                                                 & 2.23                                                 & 3.46                                                 & 2.19                                                 & 3.65                                                 & 2.77                                                 & 2.65                                                  & 3.47                                                 & 4.28                                                 & 3.42                                                 & 2.47                                                 \\
Dalle3           & 33.04 & 25.79                                                 & 30.96                                                & 27.64                                                & 31.59                                                & 28.51                                                & 28.44                                                & 30.62                                                & 14.22                                                & 29.77                                                & 26.54                                                & 31.37                                                 & 24.83                                                & 18.57                                                & 26.45                                                & 26.52                                                \\ \hline
Model            & en    & \begin{tabular}[c]{@{}c@{}}urd\\ (Arab)\end{tabular}  & \begin{tabular}[c]{@{}c@{}}kas\\ (Arab)\end{tabular} & \begin{tabular}[c]{@{}c@{}}kas\\ (Deva)\end{tabular} & \begin{tabular}[c]{@{}c@{}}mai\\ (Deva)\end{tabular} & \begin{tabular}[c]{@{}c@{}}awa\\ (Deva)\end{tabular} & \begin{tabular}[c]{@{}c@{}}bho\\ (Deva)\end{tabular} & \begin{tabular}[c]{@{}c@{}}hne\\ (Deva)\end{tabular} & \begin{tabular}[c]{@{}c@{}}mag\\ (Deva)\end{tabular} & \begin{tabular}[c]{@{}c@{}}sin\\ (Sinh)\end{tabular} & \begin{tabular}[c]{@{}c@{}}brx\\ (Deva)\end{tabular} & \begin{tabular}[c]{@{}c@{}}doi\\ (Deva)\end{tabular}  & \begin{tabular}[c]{@{}c@{}}gom\\ (Deva)\end{tabular} & \begin{tabular}[c]{@{}c@{}}sat\\ (Olck)\end{tabular} & \begin{tabular}[c]{@{}c@{}}snd\\ (Deva)\end{tabular} & \begin{tabular}[c]{@{}c@{}}mni\\ (Mtei)\end{tabular} \\ \hline
Stable Diffusion & 35.00 & 5.09                                                  & 4.61                                                 & 4.34                                                 & 4.34                                                 & 4.63                                                 & 4.22                                                 & 4.52                                                 & 4.41                                                 & 3.82                                                 & 3.90                                                 & 4.06                                                  & 4.17                                                 & 3.76                                                 & 4.16                                                 & 4.72                                                 \\
Alt Diffusion    & 32.47 & 8.17                                                  & 5.84                                                 & 6.69                                                 & 7.48                                                 & 7.44                                                 & 7.37                                                 & 7.59                                                 & 7.63                                                 & 6.16                                                 & 5.67                                                 & 6.63                                                  & 6.24                                                 & 5.84                                                 & 6.68                                                 & 5.71                                                 \\
Alt Diffusion m9 & 29.22 & 11.15                                                 & 7.80                                                 & 9.96                                                 & 11.18                                                & 12.13                                                & 11.24                                                & 12.44                                                & 12.52                                                & 11.31                                                & 4.97                                                 & 9.35                                                  & 7.59                                                 & 5.71                                                 & 9.28                                                 & 5.75                                                 \\
Midjourney       & 34.11 & 4.19                                                  & 3.15                                                 & 3.62                                                 & 3.29                                                 & 4.22                                                 & 3.30                                                 & 3.88                                                 & 3.43                                                 & 2.42                                                 & 3.32                                                 & 3.36                                                  & 3.35                                                 & 2.52                                                 & 3.47                                                 & 3.64                                                 \\
Dalle3           & 33.04 & 30.35                                                 & 23.55                                                & 27.51                                                & 28.71                                                & 27.79                                                & 30.15                                                & 29.68                                                & 29.73                                                & 22.35                                                & 11.46                                                & 29.14                                                 & 23.96                                                & 2.24                                                 & 27.91                                                & 1.54                                                 \\ \hline
\end{tabular}
\end{sidewaystable}


\begin{figure*}[t]
 \centering
   \includegraphics[width=\linewidth]{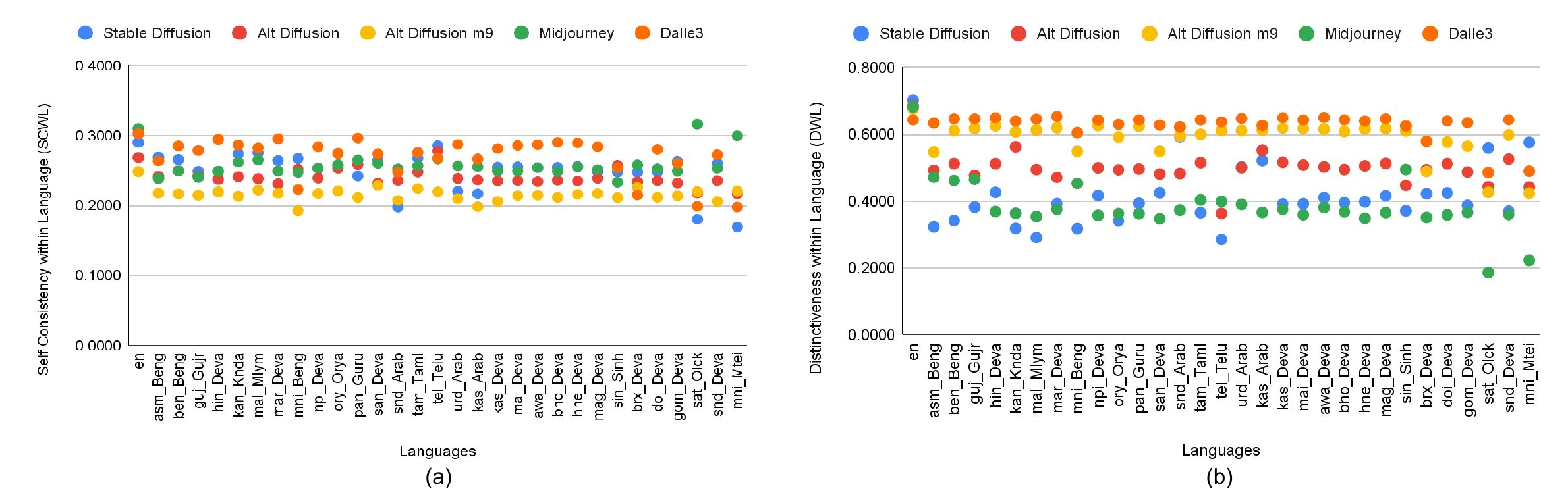}
   \caption{Performance for the (a) SCWL and (b) DWL metrics of the benchmark showcasing self-consistency and distinctiveness of concepts within language, respectively, over the complete set of prompts.}
   \label{fig:represults_supple}
\end{figure*}

\section{Benchmark Results and Analysis}
In this section, we report extended results on the common set and complete set of prompts.  The observations are consistent with those reported in the main paper on the common subset.

In the \textbf{correctness-based metrics}, the CLGC, IGC, and LGC metrics over the common set of prompts are reported in Tables \ref{tab:clgc40_2}, \ref{tab:igc40_2}, and \ref{tab:lgc40_2}, respectively. Similarly, the results for the three metrics on the complete set of prompts are reported in Tables \ref{tab:clgc40_supple}, \ref{tab:igc40_supple}, and \ref{tab:lgc40_supple}. As observed in the main paper, across all the metrics, Dalle3 outperforms all other models when evaluated for Indic languages with a significant margin. Other models struggle to generalize on the Indic languages with AltDiffusion m9 performing the best among the other three models. This behavior is a result of increased generalizability due to the multilingual training of the model in 9 languages. On the other hand, while all models perform similarly for the English language, AltDiffusion m9 performs the worst, possibly due to catastrophic forgetting when trained for multilinguality.

\begin{figure*}[t]
 \centering
   \includegraphics[width=\linewidth]{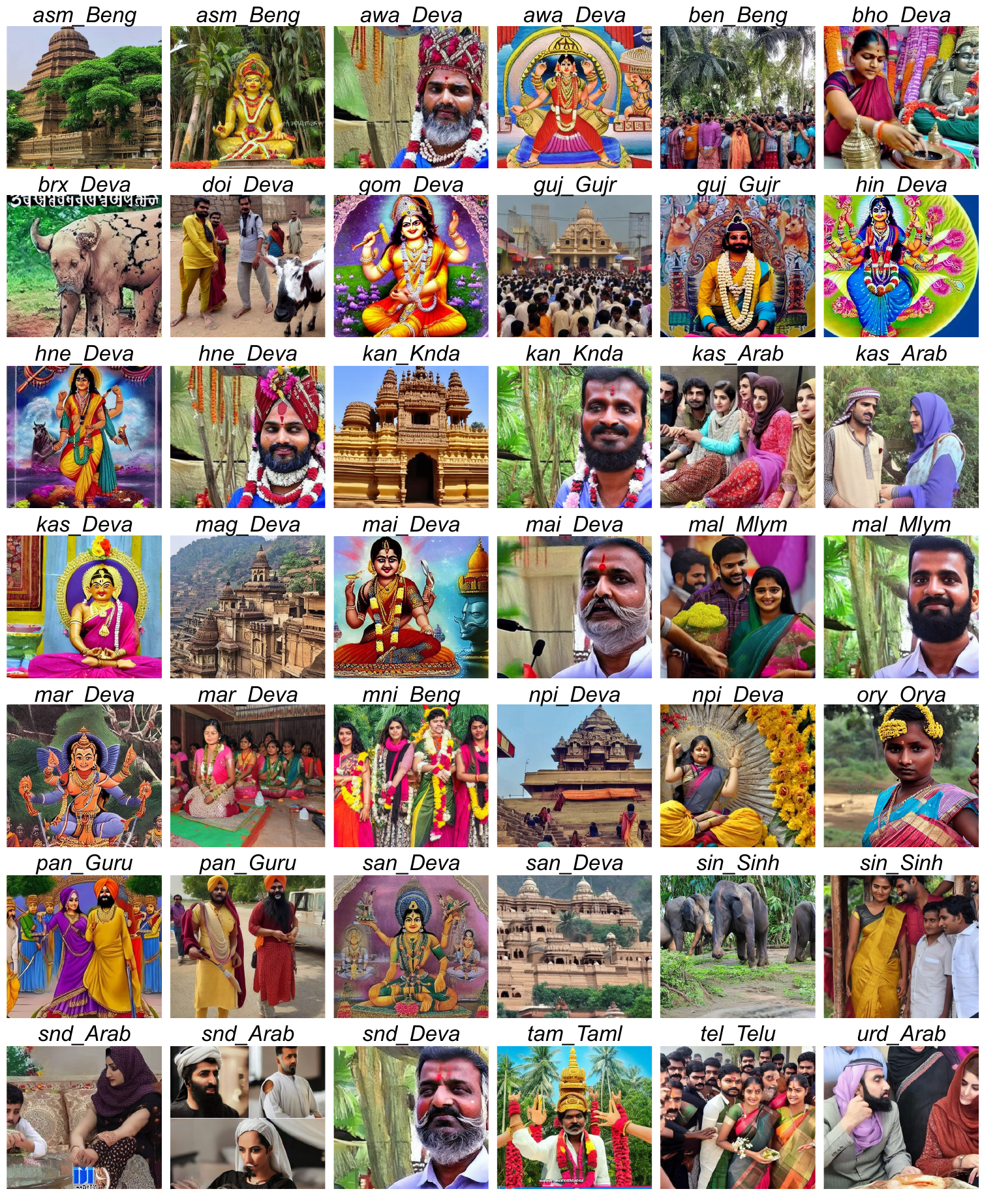}
   \caption{Showcasing the influence of language scripts on the cultural aspects depicted in the Stable Diffusion model.}
   \label{fig:sd_culture}
\end{figure*}

In \textbf{representation-based metrics}, we evaluate the SCWL and DWL metrics on the complete set and observe that they follow the same patterns as their performance on the common set (refer Fig. \ref{fig:represults_supple}). It is also observed that while AltDiffusion m9 has high distinctiveness across the concepts it generates, it provides poor self-consistency within the language, highlighting its instability and a tendency to seemingly generate diversely with or without relevance to the prompt. For the SCAL metric, the value obtained for AltDiffusion m9 on the common set of prompts comes out to be 21.47\%, which is lower than the overall \textit{SCAL metric} for the Stable Diffusion, AltDiffusion, Midjourney, and Dalle3 models is 25.44\%, 23.73\%, 26.75\%, and 29.90\%, respectively. This indicates an overall low consistency of AltDiffusion m9 in generating concepts across different languages.

\section{Qualitative Analysis}
\begin{figure*}[t]
 \centering
   \includegraphics[width=\linewidth]{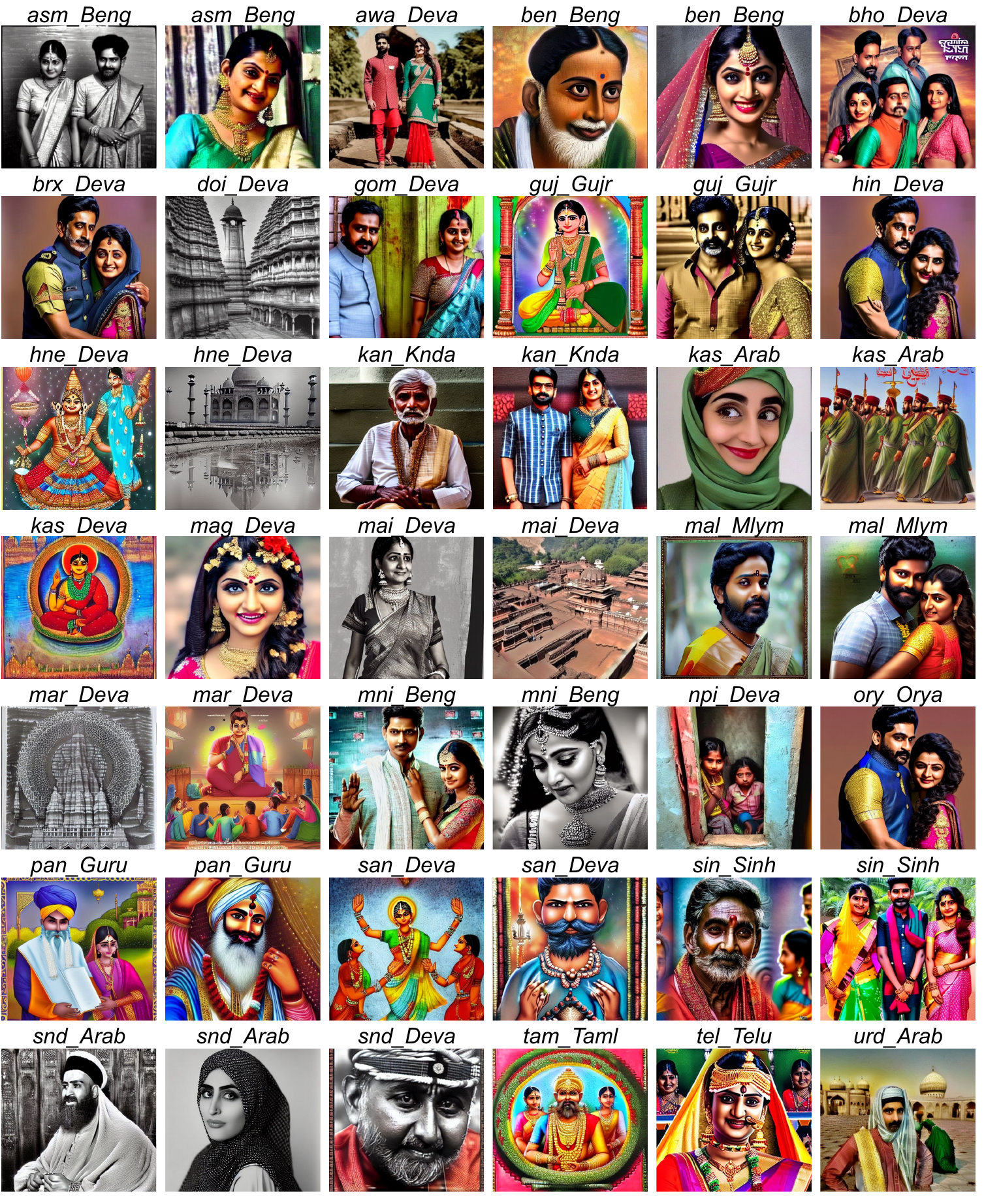}
   \caption{Showcasing the influence of language scripts on the cultural aspects depicted in the Alt Diffusion model.}
   \label{fig:altd_culture}
\end{figure*}

\begin{figure*}[t]
 \centering
   \includegraphics[width=\linewidth]{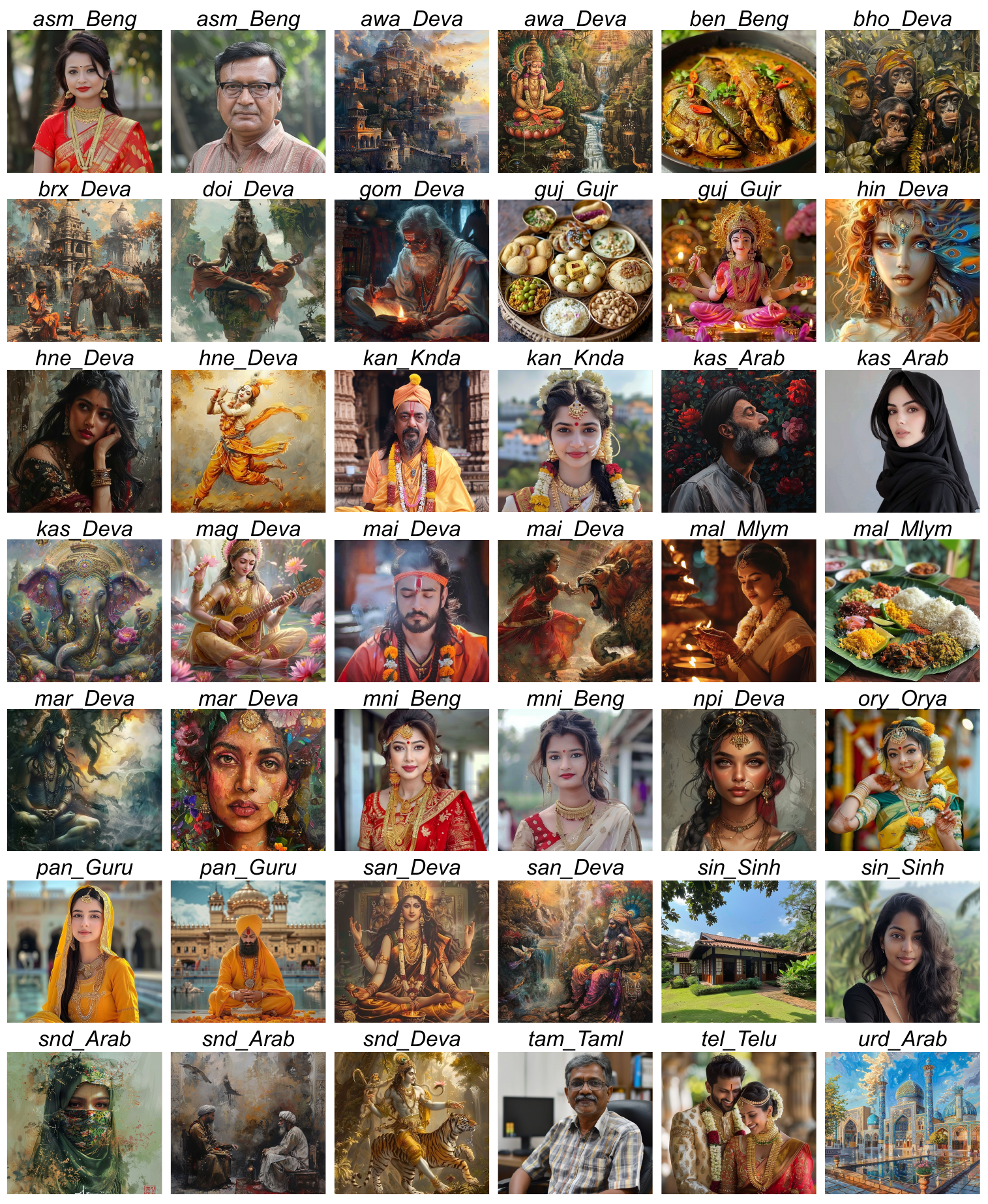}
   \caption{Showcasing the influence of language scripts on the cultural aspects depicted in Midjourney.}
   \label{fig:mid_culture}
\end{figure*}

In this main paper, we qualitatively analyzed the images generated by the different models across the different languages. We present more qualitative results here in Figs. \ref{fig:sd_culture}, \ref{fig:altd_culture}, and \ref{fig:mid_culture} corresponding to Stable Diffusion, Alt Diffusion, and Midjourney, respectively.

The Stable Diffusion model (Fig. \ref{fig:sd_culture}) generates images containing a high number of individuals, temples, flowers, and gods. In the case of Arabic scripts (kas\_Arab, snd\_Arab, urd\_Arab), the model produces men wearing Muslim caps and women in burkhas or niqabs. For all Devanagri scripts as well as for as guj\_Gujr in Gujarati script, the model generates individuals with sarees, and \textit{tilak} which are often worn in the Indian culture. Additionally, it produces gods and temples. Sanskrit (sans\_Deva) in particular, generates a large number of gods and temples due to its extensive religious context. This pattern is present across all languages using the Devanagri script. For languages using the Bengali script (asm\_Beng, ben\_Beng, mni\_Beng), the model produces a significant amount of distinctive greenery. Sindhi in Arabic script (snd\_Arab) is the only language generating a substantial amount of pornographic content. 

The AltDiffusion model (Fig. \ref{fig:altd_culture}) generates images of couples, gods, places of worship, and occasionally of Indian monuments such as the Taj Mahal, especially for languages with the Devanagari script. When generating for Arabic scripts, such as kas\_Arab, snd\_Arab, and urd\_Arab, the model generates images of men wearing Muslim caps, women in burkhas or niqabs, and mosques showcasing a correlation between the Arabic script and Muslim culture. In the Gurumukhi script (pan\_Guru), the model depicts individuals with long beards and turbans, highlighting correlations between the script of the Punjabi language with stereotypical portrayal of people from Punjab. Languages from the Dravidian family, including Telugu (tel\_Telu), Tamil (tam\_Taml), Malayalam (mal\_Mlym), and Kannada (kan\_Knda), along with Sinhala (sin\_Sinh), feature images of dark-skinned individuals, possibly associating skin-color with the individuals being generated.

\begin{figure*}[t]
 \centering
   \includegraphics[width=\linewidth]{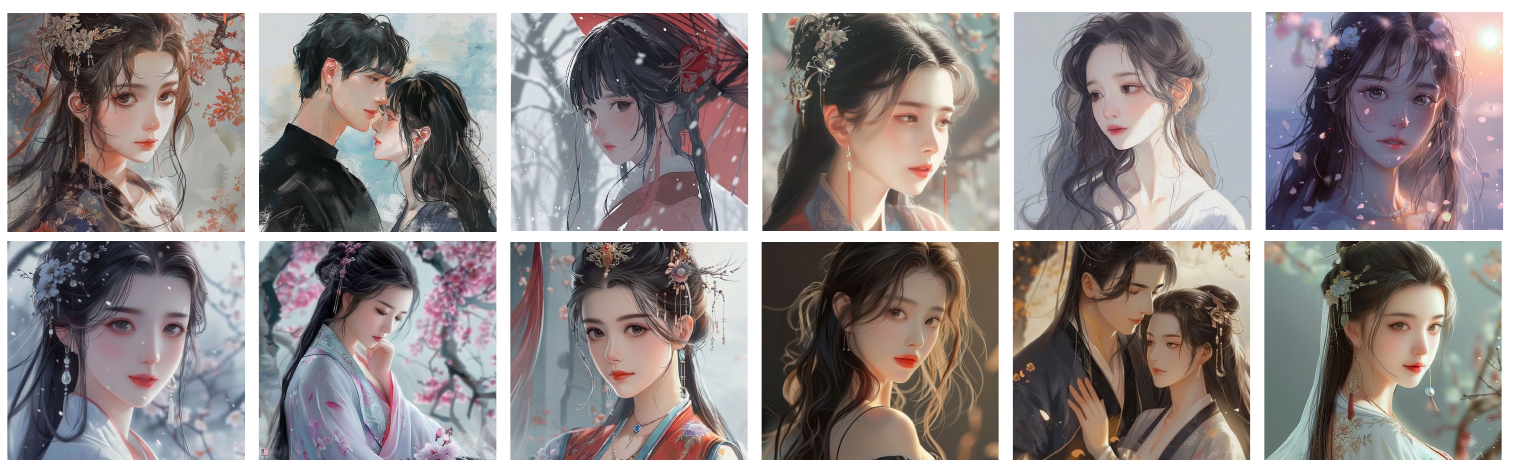}
   \caption{Showcasing the random generation of anime-style women and men in Midjourney when prompted to generate in Santali (Olck script) and Manipuri (Meitei script).}
   \label{fig:midrandom}
\end{figure*}

In Midjourney (Fig. \ref{fig:mid_culture}), for the Devanagari script, the model produces a variety of visuals, including women, gods, and deities in Hinduism (like \textit{Shiva}, \textit{Ganesha}, and \textit{Krishna}), \textit{pandits} (priests), temple-like structures (religious places of worship), elephants, and tigers (typically shown in stereotypical depictions of India). Within the Dravidian family languages such as Malayalam (mal\_Mlym), Kannada (kan\_Knda), Telugu (tel\_Telu), and Tamil (tam\_Taml), common elements include jewelry such as necklaces, forehead pendants, earrings, and dark-skinned individuals, particularly men. Bengali scripts (asm\_Beng, ben\_Beng, mni\_Beng) often feature bridal women in red sarees adorned with jewelry.  Other commonly generated images include images of food, such as fish, which is prevalent in Bengali culture. Languages with Arabic script like Kashmiri (kas\_Arab), Urdu (urd\_Arab), and Sindhi (Snd\_Arab) commonly depict women with \textit{hijabs} or \textit{niqabs}, men in headcovers, and places of Islamic worship such as mosques. In Gujarati, in addition to Devanagari influences, food-related imagery is prominent. Finally, Punjabi in Gurumukhi script (pun\_Guru) frequently showcases individuals wearing turbans with a \textit{Gurudwara} (place of worship in \textit{Sikhism}, a religion commonly practiced by residents of Punjab) in the background. For certain languages such as Santali (sat\_Olck) and Manipuri in Meitei script (mni\_Mtei), Midjourney generates Asian women, depicting no Indian cultural influences (Fig. \ref{fig:midrandom}. It is interesting to note that these two languages also produce random outputs in the Dalle3 model, which understands many of the other Indic languages. \\

\end{document}